\documentclass[]{interact}

\usepackage[caption=false]{subfig}

\usepackage[numbers,sort&compress]{natbib}
\bibpunct[, ]{[}{]}{,}{n}{,}{,}
\renewcommand\bibfont{\fontsize{10}{12}\selectfont}
\makeatletter
\def\NAT@def@citea{\def\@citea{\NAT@separator}}
\makeatother

\theoremstyle{plain}

\theoremstyle{definition}

\theoremstyle{remark}

\usepackage{amsmath}
\usepackage{url}
\usepackage{multibib}
\usepackage[normalem]{ulem}
\usepackage{appendix}
\usepackage{algorithm}
\usepackage{algorithmic}
\usepackage{amssymb}
\usepackage{amsfonts}
\usepackage{multirow}
\newcommand{\sym}[1]{\ifmmode^{#1}\else\(^{#1}\)\fi}
\usepackage{tabularx}
    
\usepackage{makecell}

\usepackage[most]{tcolorbox}
\usepackage{xcolor}
\usepackage{float}
\usepackage{cleveref}
\usepackage{etoolbox}
\tcbuselibrary{skins,breakable,theorems}

\newtcolorbox[
  auto counter,
  crefname = {Prompt.}{Prompts.}
]{prompt}[3][]
{
  enhanced,
  breakable,
  fonttitle=\bfseries,
  fontupper=\scriptsize,
  fontlower=\scriptsize,
  left=2mm,right=2mm,top=1mm,bottom=1mm,
  float,
  floatplacement=tb,
  title = Prompt.~\thetcbcounter\ifstrempty{#2}{}{~#2},
  #1,
  label = prompt:#3
}

\usepackage{listings}

\lstdefinelanguage{json}{
  basicstyle=\ttfamily\small,
  stringstyle=\color{red},
  commentstyle=\color{gray},
  morestring=[b]",
  showstringspaces=false,
  breaklines=true
}

\usepackage{type1cm}

\begin{document}

\articletype{ARTICLE TEMPLATE}

\title{Whose Is This?: Context-Aware Object Ownership Inference with Uncertainty-Guided Questioning}

\author{
\name{
Saki Hashimoto\textsuperscript{a}
and Akira Taniguchi\textsuperscript{b}\textsuperscript{*}\thanks{CONTACT Akira Taniguchi. Email: taniguchi.akira255@mail.kyutech.jp}
and Shoichi Hasegawa\textsuperscript{a}
and Yoshinobu Hagiwara\textsuperscript{c,d}
and Tadahiro Taniguchi\textsuperscript{e,d}
}
\affil{
\textsuperscript{a}Graduate School of Information Science and Engineering, Ritsumeikan University, Osaka, Japan\\
\textsuperscript{b}College of Information Science and Engineering, Ritsumeikan University, Osaka, Japan\\
\textsuperscript{*}Present affiliation is Faculty of Computer Science and Systems Engineering, Kyushu Institute of Technology, Fukuoka, Japan\\
\textsuperscript{c}Faculty of Science and Engineering, Soka University, Tokyo, Japan\\
\textsuperscript{d}Research Organization of Science and Technology, Ritsumeikan University, Shiga, Japan\\
\textsuperscript{e}Graduate School of Informatics, Kyoto University, Kyoto, Japan
}
}


\maketitle

\begin{abstract}
Service robots must infer object ownership to correctly interpret instructions such as “bring me my cup.” However, ownership is a latent attribute that cannot be directly observed, and existing methods often rely on limited cues such as recent usage, making them unreliable in scenarios such as temporary sharing.
We propose a framework for context-aware ownership inference with uncertainty-guided interaction (COIN). The method integrates user background information and object usage history using a large language model (LLM) to estimate ownership scores. To handle uncertainty, we apply conformal prediction to construct a set of plausible owners and selectively generate user queries when the prediction is uncertain.
Experiments in a simulated home environment show that the proposed method consistently outperforms baseline approaches, achieving a Subset Accuracy of 0.988 and a Mean Jaccard index of 0.991. The method also maintains high performance in scenarios involving temporary use and shared ownership. 
The results demonstrate that combining contextual reasoning with uncertainty-aware interaction improves both estimation accuracy and robustness.
The project page is available at \url{https://emergentsystemlabstudent.github.io/COIN/}.

\end{abstract}


\section{Inroduction}\label{sec:introduction}

To correctly respond to language instructions such as bring me my cup,' a service robot must identify the object specified by the user. 
In everyday environments, recognizing object ownership---determining `who owns what'---is essential for this capability. 
However, ownership is a latent attribute that cannot be directly observed, making it difficult to infer from visual features or spatial information alone.

A variety of methods have been proposed to estimate object ownership by leveraging interactions between users and objects. 
For example, some approaches rely on observable attributes such as color, location, and the most recent user~\cite{tan2019s,ActOwL}, while others use interaction histories derived from human-object relationships such as `carrying' or `holding'~\cite{wu2020item,hu2023interactive}. 
These methods demonstrate that interaction provides useful cues for ownership estimation. However, they primarily rely on observable behavioral signals, making it difficult to distinguish ownership from temporary use. 
For instance, even if a child is using a computer, the device may belong to another family member and be borrowed only temporarily. 
In addition, these approaches do not sufficiently exploit richer contextual information, such as user background and long-term usage patterns.

Even when contextual information is incorporated, ownership estimation remains uncertain in real-world environments. 
In such cases, acquiring additional information through user interaction is beneficial; however, querying users about every object imposes excessive burden. 
Therefore, it is necessary to selectively generate queries based on estimation uncertainty.

\begin{figure}[tb]
    \centering
    \includegraphics[width=0.9\linewidth]{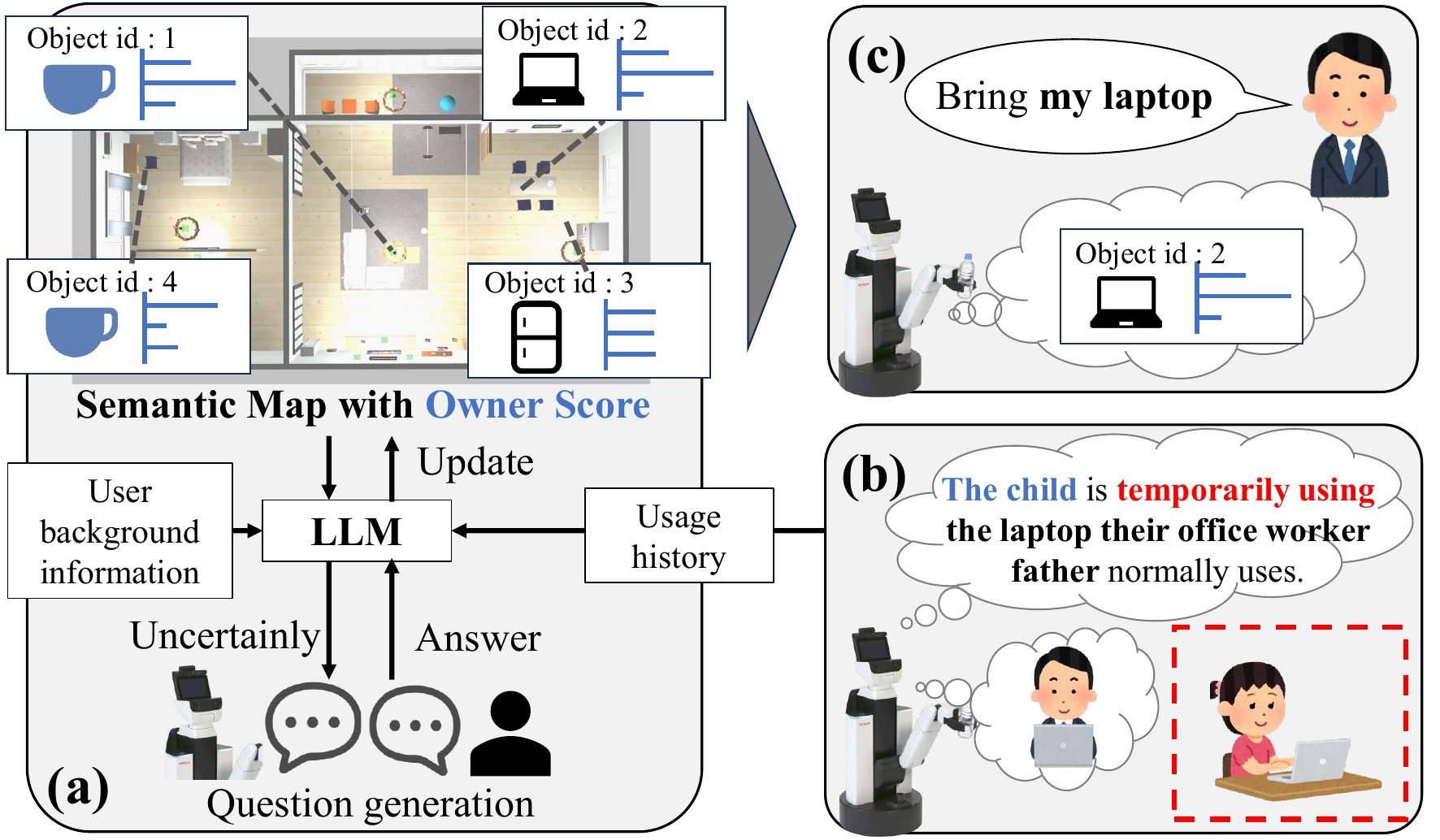}
    \caption{Overview of the proposed framework.
    (a) The robot estimates ownership scores using an LLM based on user background information and object usage history. 
    When the estimation is uncertain, it prompts the user to confirm object ownership. 
    The responses are used to update the ownership scores, which are then stored in the semantic map.
    (b) The object usage history captures differences in users and their interactions with objects, enabling the robot to recognize temporary usage.
    (c) By leveraging the semantic map augmented with ownership scores, the robot can interpret language instructions that include ownership references.
    }
    \label{fig:research_overview}
\end{figure}

In this paper, we propose Context-aware Ownership inference with INteraction (COIN), a framework for ownership estimation that integrates contextual reasoning with uncertainty-aware interaction. 
The proposed method extends NLMap~\cite{NLMap} by introducing an ownership-aware representation that assigns ownership scores to candidate users. 
Ownership is estimated using a large language model (LLM) that incorporates user background information and object usage history, while uncertainty is quantified using conformal prediction (CP)~\cite{CP2005}. 
Based on this uncertainty, the robot selectively generates questions to acquire additional information. An overview of the proposed framework is shown in Figure~\ref{fig:research_overview}.

The proposed approach aims to accurately estimate object ownership while minimizing interaction cost. 
By integrating contextual information and selectively acquiring additional information based on uncertainty, the method balances estimation accuracy and interaction efficiency.

To evaluate the proposed method, we conduct experiments in a simulated home environment and compare its performance with existing approaches. The evaluation focuses on both ownership estimation accuracy and the number of user queries. We also perform ablation studies to analyze the contribution of key components, including contextual information, uncertainty estimation, and question generation.

The main contributions of this work are as follows:
\begin{enumerate}
    \item Integration of contextual information, including user background and usage history, for ownership estimation beyond observable cues
    \item An uncertainty-aware interaction strategy that reduces unnecessary queries while maintaining high estimation accuracy
\end{enumerate}

The remainder of this paper is organized as follows. 
Section~\ref{sec:problem_statement} formulates the problem addressed in this study. 
Section~\ref{sec:related_work} reviews related work. 
Section~\ref{sec:proposed_method} presents the proposed method. Section~\ref{sec:experiment1} describes comparative experiments with existing methods. 
Section~\ref{sec:experiment2} presents ablation studies. Section~\ref{sec:limitation} discusses the limitations of this work. Finally, Section~\ref{sec:conclusion} concludes the paper and outlines future work.

\section{Problem Statement}\label{sec:problem_statement}

Existing studies~\cite{tan2019s,wu2020item,hu2023interactive,ActOwL} on object ownership estimation exhibit the following three major limitations.

\textbf{Problem 1: Limited use of contextual information.}  
A primary limitation is the insufficient use of rich contextual information about users. Existing approaches~\cite{hu2023interactive,wu2020item,tan2019s,ActOwL} mainly rely on observable cues, such as spatial proximity, visual features, and limited interaction histories, to estimate ownership. 
However, in real-world environments, higher-level contextual information--such as user roles (\textit{e.g.}, parent or child), occupations, daily routines, and preferences--often provides crucial clues for identifying object ownership. The inability to effectively incorporate such contextual knowledge significantly constrains the performance of existing methods.

\textbf{Problem 2: Inability to handle dynamic and temporary usage.}  
A second limitation is the assumption that object ownership is static or temporally stable. Existing methods~\cite{hu2023interactive,wu2020item,tan2019s,ActOwL} typically do not account for dynamic situations in which objects are temporarily used, moved, or borrowed by others. 
In real-world environments, such situations frequently occur, and it is essential to distinguish between the current user of an object and its actual owner. However, existing approaches do not explicitly model these transient usage patterns.

\textbf{Problem 3: Lack of uncertainty-aware interaction.}  
Another important limitation is the absence of mechanisms to handle uncertainty in ownership estimation. Existing methods do not explicitly quantify estimation uncertainty or adapt their behavior accordingly. 
As a result, they cannot selectively acquire additional information when needed, often leading to either insufficient information or unnecessary user queries.

\section{Related Work}\label{sec:related_work}

This section reviews prior studies related to object ownership estimation in everyday environments and clarifies the positioning of this work. 
Section~\ref{sec:rw_ownership} provides an overview of existing ownership estimation methods. 
Section~\ref{sec:rw_psych} discusses findings from psychological studies on ownership. 
Finally, Section~\ref{sec:rw_uncertainty} reviews research on uncertainty representation and question generation.


\subsection{Ownership Inference in Robotics}
\label{sec:rw_ownership}
For robots to handle instructions that include ownership references in everyday environments, it is essential to learn and continuously update the associations between objects and users. 
Tan~\textit{et~al.}~\cite{tan2019s} proposed a method that represents ownership relationships probabilistically and updates them through user feedback. 
Wu~\textit{et~al.}~\cite{wu2020item} estimated object ownership based on interaction histories between humans and objects.
Hu~\textit{et~al.}~\cite{hu2023interactive} enhanced robustness in real-world environments by integrating interactive learning, incorporating user corrections, into an autonomous learning framework based on human–object interaction detection and person re-identification. 
ActOwL~\cite{ActOwL} proposed a framework that learns ownership relationships by combining commonsense knowledge from LLMs with probabilistic generative models, and actively acquires knowledge by generating questions that maximize information gain.


These approaches primarily estimate object ownership based on observed interactions. 
However, in everyday environments, the current user of an object does not always correspond to its actual owner. 
In such cases, relying solely on interaction cues---such as physical contact or recent use---is often insufficient for accurate ownership identification. 
In contrast, our work integrates richer contextual information, including user background and object usage history, together with spatial and attribute information.


\subsection{Psychological Ownership and Context Modeling}
\label{sec:rw_psych}
In psychology and consumer behavior research, the concept of \emph{psychological ownership} refers to a state in which individuals subjectively feel that an object is `theirs,' independent of legal ownership~\cite{morewedge2021evolution,iseki2022development}. 
Key factors contributing to this perception include a sense of control over the object, knowledge about the object, and personal investment in it. In addition, physical contact with an object has been shown to increase psychological ownership~\cite{peck2009effect}. 
Iseki~\textit{et~al.} developed a psychological ownership scale (POS-J) and demonstrated that the sense of ownership can vary continuously depending on situational context and the level of user involvement~\cite{iseki2022development}.

These findings suggest that interactions with objects, such as contact and usage experience, influence the perception that `this object belongs to me.' 
In other words, ownership is not merely a fixed label, but is shaped by the history of engagement between a user and an object. 


\subsection{Uncertainty Modeling and Interactive Information Acquisition}
\label{sec:rw_uncertainty}
Language instructions from users in everyday environments often contain ambiguity, and environment-specific information, such as object ownership, cannot be directly observed by robots. 
Therefore, robots must explicitly account for uncertainty in their estimates and acquire additional information when necessary.

Existing approaches to handling uncertainty can be broadly categorized into two groups. 
The first focuses on calibrating and presenting the reliability of predictions. 
For example, some methods present candidate objects as ranked lists~\cite{MultiRankIt}, while others use Conformal Prediction~(CP)~\cite{CP2005} to construct prediction sets and control whether confirmation is required~\cite{Lap,IntroPlan,KNOWNO}. 
These approaches aim to reduce erroneous actions caused by overconfidence by explicitly representing uncertainty.

The second group focuses on actively acquiring information by generating questions when uncertainty is high. 
This includes frameworks that iteratively generate questions for reference resolution or personalized goal search~\cite{ORION,oyama_take_2025}, methods that integrate questioning into action policies~\cite{AUTOASK}, and approaches that optimize questions based on information gain~\cite{apricot,UoT}. 
These methods emphasize active information acquisition to reduce uncertainty.

In addition, recent studies have explored planning and exploration methods using foundation models to address broader sources of uncertainty, such as observation noise and ambiguity in language understanding~\cite{Navid,OpenFMNav,SPINE}.

In contrast, this study focuses on uncertainty specific to ownership estimation, \textit{i.e.}, determining `who owns an object.' 
Even when reasoning from observations and usage history, insufficient evidence may result in multiple plausible owner candidates. 
To address this, we do not directly interpret LLM output scores as probabilities; instead, we construct prediction sets of candidate owners using CP~\cite{CP2005}. 
The size of the prediction set is used as an uncertainty measure, and questions are generated only when the set is large. 
This enables efficient information acquisition while maintaining statistical guarantees.

\section{Proposed Method: Context-aware Ownership inference with INteraction (COIN)}\label{sec:proposed_method}

\begin{figure}[tb]
    \centering
    \includegraphics[width=1.0\linewidth]{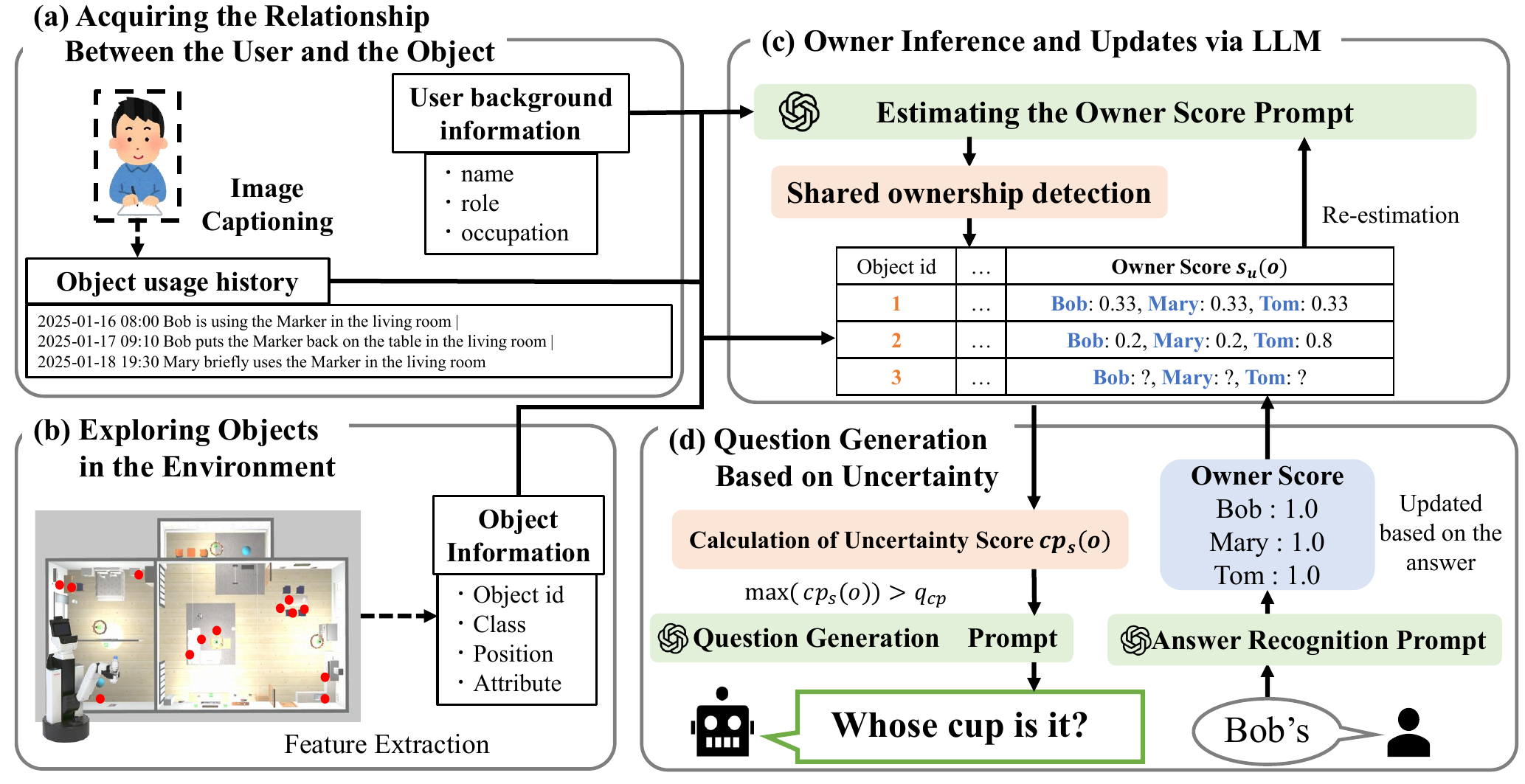}
    \caption{Overview of the proposed method.
    (a) The robot is assumed to have access to user background information obtained in advance. It also collects and accumulates interaction histories between users and objects based on observations (see Section~\ref{subsec:env_info}).
    (b) The robot explores the environment and acquires object-level information, including spatial locations, object classes, and visual features (see Section~\ref{subsec:object_exploration}).
    (c) By integrating user background, object usage history, and observed object information, the robot estimates ownership scores for each object. The estimates are iteratively updated, allowing the system to handle temporary usage and borrowing scenarios (see Section~\ref{subsec:llm_inference}).
    (d) The uncertainty of the ownership estimation is evaluated, and the robot selectively generates questions only for objects with high uncertainty. This reduces unnecessary queries while improving estimation accuracy (see Section~\ref{subsec:question_generation}).
    }
    \label{fig:proposed_method}
\end{figure}

We propose a method for estimating object ownership in everyday environments by leveraging user background information and object usage history. 
An overview of the proposed framework is illustrated in Figure~\ref{fig:proposed_method}, and the algorithm is summarized in the Appendix~\ref{app:algorithm}.

The proposed method iteratively estimates ownership by combining context-based inference with uncertainty-aware interaction. 
First, the robot constructs a set of reliable known information from the current ownership estimation state. 
Then, it estimates ownership scores for each candidate user using an LLM based on object attributes and contextual information, including usage history. 
Next, the estimation uncertainty is quantified using CP. 
If the uncertainty is high, the robot generates a question for the user to acquire additional information. 
The user's response is then interpreted and incorporated into the ownership estimation state, and the estimation is updated accordingly. 
This process is repeated until a confident ownership estimation is obtained.

\subsection{Acquiring User–Object Relationship Information}\label{subsec:env_info}

To estimate object ownership, our method leverages user background information and previously collected object usage history. 
These sources of information serve as contextual inputs to the LLM for ownership inference.

\subsubsection{User Background Information}

We assume that user background information is given in advance and define its structured representation. This information provides high-level contextual cues for ownership estimation. Each user is associated with the following attributes:

\begin{itemize}
    \item \texttt{name}: identifier of the user
    \item \texttt{role}: role in the environment (\textit{e.g.}, father, mother, child)
    \item \texttt{occupation}: occupation or primary activity
\end{itemize}

The background information is represented as structured text and provided as part of the LLM's input context. An example representation is shown below.

\begin{lstlisting}
{
  "Bob": {
    "role": "father",
    "occupation": "office worker"
  },
  "Mary": {
    "role": "mother",
    "occupation": "homemaker"
  },
  "Tom": {
    "role": "son",
    "occupation": "elementary school student"
  }
}
\end{lstlisting}

By incorporating such background information, the model can perform ownership estimation that reflects user roles and contextual knowledge, rather than relying solely on spatial or visual cues.

\subsubsection{Acquisition of Usage History}\label{subsubsec:usage_history}

To capture interactions between users and objects, we assume that action descriptions are generated from environmental observations using a vision-language model (VLM) via image captioning. Each caption describes the timestamp, actor, target object, and action in natural language.

Examples of such captions are shown below:

\begin{lstlisting}
2025-01-19 19:00 Bob takes the Marker
2025-01-19 19:00 Bob puts the Marker back
2025-01-20 10:00 Mary looks for the Marker
\end{lstlisting}

These captions are stored together with timestamps and accumulated as usage history for each object. At this stage, the captions are preserved in their original natural language form, without explicit structuring or conversion into predefined action labels.

Storing interaction histories in natural language enables the system to retain the richness and context-dependency of human behavior. These histories are later interpreted as needed during the ownership inference process.

\subsection{Exploring Objects in the Environment}\label{subsec:object_exploration}

To systematically manage objects in the environment, we adopt an extended version of NLMap~\cite{NLMap}, an open-vocabulary semantic map that associates natural language queries with object instances using visual and geometric representations. This enables the robot to consistently retrieve candidate objects corresponding to expressions such as `my cup.' A detailed formulation of NLMap is provided in Appendix~\ref{app:NLMap}.

In this study, we extend NLMap to better support object management in everyday environments. The main modifications are as follows:
\begin{enumerate}
    \item We replace the object detector ViLD~\cite{ViLD} with Detic~\cite{Detic} to improve detection performance, particularly for small objects.
    \item We introduce an instance-level integration mechanism that assigns consistent object IDs across multiple observations.
    \item We extend the spatial representation from 2D to 3D by modeling object locations with 3D Gaussian distributions.
\end{enumerate}

These extensions enable robust and consistent object representation by jointly modeling semantic and spatial information. The system can reliably detect diverse objects, integrate observations from multiple viewpoints into a single object instance, and reduce erroneous associations between nearby objects by incorporating height information.

Furthermore, we augment the extended NLMap with ownership estimation, constructing an \emph{ownership-aware NLMap}. 
For each object instance $o$, the map maintains ownership scores $\{ s_u(o) \mid u \in \mathcal{U} \}$ over candidate users. 
The ownership score represents a heuristic confidence level rather than a calibrated probability.
Each object is represented by the following attributes:
\begin{itemize}
    \item object ID
    \item object class label
    \item 3D spatial location
    \item visual features (CLIP embeddings)
    \item ownership scores $\{ s_u(o) \}$
\end{itemize}

This unified representation enables the system to jointly manage object properties and uncertain ownership information. It also facilitates consistent ownership inference and updates at the object level, allowing the robot to generate appropriate queries and directly incorporate user feedback into the corresponding ownership scores.






\subsection{Owner Inference and Updates via LLM}\label{subsec:llm_inference}

We estimate object ownership using an LLM that integrates multiple sources of contextual information. Given a target object $o$, the LLM takes as input a prompt constructed from user background information, object usage history, and object-level attributes obtained from the semantic map.

The input to the LLM consists of the following components:
\begin{itemize}
    \item \textbf{User background}: contextual attributes such as roles and occupations
    \item \textbf{Usage history}: natural language action descriptions generated by a VLM
    \item \textbf{Object information}: class label, spatial location, and visual features stored in the NLMap
\end{itemize}

Based on these inputs, the LLM estimates ownership scores $\{ s_u(o) \mid u \in \mathcal{U} \}$ for each candidate user. To accommodate ambiguous ownership and shared-use scenarios, the scores are not normalized and do not necessarily sum to 1, forming a multi-label representation.

\subsubsection{Spatially-Aware Context Extraction}\label{subsubsec:spatial_context}

We incorporate spatial context by extracting objects in the vicinity of the target object. Let $(x_o, y_o, z_o)$ denote the position of object $o$, and $(x_i, y_i, z_i)$ denote the position of another object $i$. The distance $d_i$ is defined as:
\begin{align}
d_i = \sqrt{(x_i - x_o)^2 + (y_i - y_o)^2 + (\gamma (z_i - z_o))^2}
\end{align}
where $\gamma$ is a scaling factor that adjusts the contribution of the vertical axis.

To obtain a smooth measure of spatial proximity, we compute a Gaussian weight:
\begin{align}
w_i = \exp\left(-\frac{d_i^2}{2\sigma^2}\right)
\end{align}
where $\sigma$ controls the spatial neighborhood range.

Objects with weights below a threshold are discarded, and the remaining candidates are ranked by $w_i$ (and distance as a tie-breaker). 
The top-$K_{\mathrm{near}}$ objects are selected as neighboring objects and formatted as structured input to the LLM, including object ID, class label, distance, and weight. 
If ownership estimates for neighboring objects are available, they are also included as contextual cues.

\subsubsection{Visually Similar Object Retrieval}\label{subsubsec:visual_similarity}

We incorporate visual context by retrieving objects with a similar appearance to the target object. Objects with similar visual features are likely to share ownership patterns, particularly in environments where users exhibit consistent preferences.

Visual similarity is computed using feature embeddings stored in the NLMap. Each object is represented by a 512-dimensional feature vector extracted by the CLIP image encoder. Let $\hat{\mathbf{f}}_o$ and $\hat{\mathbf{f}}_i$ denote the L2-normalized feature vectors of the target object $o$ and a candidate object $i$, respectively. The similarity is defined as the cosine similarity:
\begin{align}
\mathrm{sim}(o, i) = \hat{\mathbf{f}}_o^\top \hat{\mathbf{f}}_i
\end{align}

Based on this similarity, top-$K_{\mathrm{sim}}$ objects (excluding the target object itself) are selected as visually similar objects. 
These objects are formatted as structured input to the LLM, including object ID, class label, and similarity score. 
If ownership estimates for these objects are available, they are also incorporated as contextual cues to support ownership inference.




\subsubsection{Usage History Extraction}\label{subsubsec:usage_history_extract}

Usage history provides one of the most direct cues for ownership estimation. We extract structured summaries from natural language interaction logs to capture how frequently, recently, and in what manner each user interacts with a target object.

To obtain stable and structured inputs for the LLM, we employ an external summarization module that aggregates raw action captions into user-level statistics. The module is implemented as a Model Context Protocol (MCP) server~\cite{fastmcp}, enabling the LLM to retrieve usage summaries via tool calls. Details of MCP are provided in Appendix~\ref{sec:MCP}.

Given a set of time-stamped action captions, each event is parsed into a tuple consisting of timestamp, user, and action. Actions are categorized into predefined types (\textit{e.g.}, \texttt{use}, \texttt{place}, \texttt{transport}, \texttt{clean}), and aggregated per user to compute statistics such as event counts, action distributions, and recency.

The resulting summary is provided to the LLM in a structured JSON format. An example output of the \texttt{get\_usage\_summary} function is shown below.

\begin{lstlisting}
{
  "object": { "id": "3", "name": "Marker" },
  "user_summary": [
    {
      "user_id": "Bob",
      "total_events": 5,
      "actions": { "use": 4, "place": 1 },
      "last_used_days_ago": 0.8,
      "example_events": [
        "2025-01-16 08:00 Bob is using the Marker",
        "2025-01-17 09:10 Bob puts the Marker back"
      ]
    },
    {
      "user_id": "Mary",
      "total_events": 1,
      "actions": { "use": 1 },
      "last_used_days_ago": 12.4,
      "example_events": [
        "2025-01-18 19:30 Mary briefly uses the Marker"
      ]
    }
  ]
}
\end{lstlisting}

By leveraging such structured usage summaries, the model can capture user-specific interaction patterns, enabling ownership estimation that reflects both long-term tendencies and recent usage.

\subsubsection{Ownership Score Estimation and Update}

We estimate ownership scores by integrating multiple sources of information into a single prompt for the LLM. 
The input includes object attributes (ID and class), spatially neighboring objects $\mathcal{N}(o)$, visually similar objects $\mathcal{S}(o)$, and usage history summaries $\mathcal{U}(o)$.

Including the object class enables the LLM to leverage commonsense knowledge alongside user background information. 
In addition, if ownership scores for related objects are already available, they are incorporated as contextual cues to capture local ownership patterns.

\paragraph{Multi-label Ownership Score Estimation via LLM}

Given the integrated prompt (see Prompt~\ref{prompt:prompt_single}), the LLM estimates ownership scores for a target object $o$ over a set of candidate users $\mathcal{U}$. Each score is independently predicted as:
\begin{align}
\mathbf{s}(o) = \{ s_u(o) \mid u \in \mathcal{U} \}, \quad s_u(o) \in [0,1]
\end{align}

The LLM is prompted to produce multi-label scores without enforcing normalization, allowing the representation of shared ownership and ambiguous cases. 
This formulation enables flexible ownership inference, allowing multiple users to be associated with a single object.



\begin{prompt}{Ownership Inference Prompt}{prompt_single}
1: You are an excellent household robot.\\
2: Your task: estimate ownership probability for ONE target object using ALL given information sources.\\
3:\\
4: IMPORTANT:\\
5: - This is MULTI-LABEL probability (each person independently between 0 and 1).\\
6: - The probabilities MUST NOT sum to 1.\\
7: - Output numbers only (0..1), no formulas.\\
8:\\
9: [\textcolor{teal}{\textbf{MEMBER\_BACKGROUND}}]\\
10:\\
11: \#\#\# Target Object \\
12: - object\_id: [\textcolor{teal}{\textbf{OBJECT\_ID}}]\\
13: - class: [\textcolor{teal}{\textbf{OBJECT\_CLASS}}]\\
14:\\
15: \#\#\# Similar Objects (may include known\_ownership for asked==1 or high-confidence)\\
16: [\textcolor{teal}{\textbf{SIMILAR\_OBJECTS}}]\\
17:\\
18: \#\#\# Nearby Objects (may include known\_ownership)\\
19: [\textcolor{teal}{\textbf{NEARBY\_OBJECTS}}]\\
20:\\
21: \#\#\# Usage History (last 365 days)\\
22: [\textcolor{teal}{\textbf{USAGE\_HISTORY}}]\\
23:\\
24: \#\#\# Output \\
25: Return ONLY the following JSON (no other explanation, no code fences):\\
26: \{\{\\
27: \ \ "ownership\_distribution": \{\{"Bob":0.xx, "Mary":0.xx, "Tom":0.xx\}\}\\
28: \}\}
\tcblower
\textcolor{teal}{\textbf{MEMBER\_BACKGROUND}}: Background information for each user in the environment\\
\textcolor{teal}{\textbf{OBJECT\_ID}}: Identifier of the target object\\
\textcolor{teal}{\textbf{OBJECT\_CLASS}}: Object class of the target object\\
\textcolor{teal}{\textbf{SIMILAR\_OBJECTS}}: Information about similar objects (JSON format; may include known ownership information)\\
\textcolor{teal}{\textbf{NEARBY\_OBJECTS}}: Information about nearby objects (JSON format; may include known ownership information)\\
\textcolor{teal}{\textbf{USAGE\_HISTORY}}: Summary of past usage events (JSON format)
\end{prompt}


\paragraph{Shared Ownership Detection and Score Refinement}

In everyday environments, objects are often jointly used or managed by multiple users. 
To capture such shared ownership, we analyze the distribution of ownership scores $\mathbf{s}(o)$.

Let the scores be sorted in descending order as
\begin{align}
s_{(1)}(o) \ge s_{(2)}(o) \ge \cdots \ge s_{(|\mathcal{U}|)}(o).
\end{align}
We determine that the object $o$ is shared among the top-$k$ users, \textit{i.e.}, the owner set consists of the top-$k$ users ranked by $s_u(o)$, if the following conditions are satisfied:
\begin{align}
s_{(k)}(o) &\ge \varepsilon_{\mathrm{min}}, \\
s_{(1)}(o) - s_{(k)}(o) &\le \varepsilon_{\mathrm{in}}, \label{eq:share_in} \\
s_{(k)}(o) - s_{(k+1)}(o) &\ge \varepsilon_{\mathrm{out}}, \label{eq:share_out}
\end{align}
where $\varepsilon_{\mathrm{min}}$ denotes the minimum confidence required for a user to be considered a shared owner, 
$\varepsilon_{\mathrm{in}}$ controls the allowable variation among shared users, and $\varepsilon_{\mathrm{out}}$ enforces separation between shared and non-shared users.

The shared ownership label is stored as an attribute in the NLMap and is used in subsequent inference and update steps. 
The ownership scores $\mathbf{s}(o)$ are also maintained as the underlying representation, enabling consistent integration with other object attributes such as spatial and visual features.




\subsection{Question Generation Based on Uncertainty}\label{subsec:question_generation}

We formulate object ownership estimation as a scoring problem, where the model assigns a confidence score
$s_u(o) \in [0,1]$
to each user $u \in \mathcal{U}$ for a given object $o$.

Selecting the user with the highest score alone does not guarantee the reliability of the prediction. 
To address this limitation, we introduce Conformal Prediction (CP), which constructs a prediction set that contains the true owner with a desired confidence level.

Using CP, we obtain a prediction set $\Gamma(o) \subseteq \mathcal{U}$ for each object $o$, which represents the set of plausible owners. 
The size of this set serves as an uncertainty measure: a larger set indicates higher uncertainty in the ownership estimation.

Based on this uncertainty, the system selectively generates questions for objects with large prediction sets, while skipping queries for objects with confident predictions. 
This enables both uncertainty-aware information acquisition and principled stopping criteria.

CP provides a statistical guarantee that the true owner is included in the prediction set with a predefined confidence level, thereby providing a principled uncertainty measure to guide interaction.



\subsubsection{Calibration Phase}

In Conformal Prediction (CP), a separate calibration dataset is used to estimate the distribution of nonconformity scores, which is then reused at inference time. We split the available data into training and calibration sets, and perform the following procedure on the calibration data. Further details are provided in Appendix~\ref{sec:CP}.

\paragraph{Nonconformity Score}

We define a nonconformity score that accounts for multiple possible true owners. Let $U_{\mathrm{true}} \subseteq \mathcal{U}$ denote the set of true owners for object $o$, and let $s_u(o) \in [0,1]$ be the predicted ownership score for user $u$. The nonconformity score is defined as:
\begin{equation}
\mathrm{nc}(o, U_{\mathrm{true}})
=
1 - \max_{u \in U_{\mathrm{true}}} s_u(o).
\end{equation}

This score reflects how poorly the model assigns confidence to the true owners. It becomes small when at least one true owner receives a high score, and large otherwise.

\paragraph{Calibration via Split Conformal Prediction}

Let $N$ be the number of calibration samples, and let $\{ \mathrm{nc}_i \}_{i=1}^{N}$ denote the corresponding nonconformity scores. Following split conformal prediction, we compute the quantile threshold $q_\alpha$ as:
\begin{equation}
q_\alpha
=
\mathrm{Quantile}_{\lceil (N+1)(1-\alpha) \rceil}
\left( \{ \mathrm{nc}_i \}_{i=1}^{N} \right).
\end{equation}

Here, $\lceil \cdot \rceil$ denotes the ceiling operator. 
We set $\alpha = 0.2$, which guarantees that the true owner is included in the prediction set with probability at least $80\%$.

The resulting threshold $q_\alpha$ is fixed after calibration and used during inference.





\subsubsection{Inference Phase}

Using the calibrated threshold $q_\alpha$, we perform ownership estimation and question generation for a target object $o$.

\paragraph{Prediction Set Construction}

Given $q_\alpha$, we define a confidence threshold $1 - q_\alpha$ and construct the prediction set:
\begin{equation}
\Gamma(o)
=
\left\{
u \in \mathcal{U}
\;\middle|\;
s_u(o) \ge 1 - q_\alpha
\right\}.
\end{equation}

By the guarantee of split conformal prediction, the prediction set satisfies:
\begin{equation}
\mathbb{P}
\left(
U_{\mathrm{true}} \subseteq \Gamma(o)
\right)
\ge 1 - \alpha.
\end{equation}

\paragraph{Uncertainty Score}

To guide interaction, we define an uncertainty score based on the average confidence over the prediction set:
\begin{equation}
\mathrm{cp}_s(o)
=
1
-
\frac{1}{|\Gamma(o)|}
\sum_{u \in \Gamma(o)} s_u(o).
\end{equation}

This score lies in $[0,1]$, where larger values indicate higher uncertainty.

\paragraph{Stopping Criterion}

Let $\{ \mathrm{cp}_s^{(i)} \}_{i=1}^{N}$ be the uncertainty scores on the calibration set. We compute a lower quantile threshold:
\begin{equation}
q_{\mathrm{cp}}
=
\mathrm{Quantile}_{\lceil (N+1)\alpha_{\mathrm{cp}} \rceil}
\left(
\{ \mathrm{cp}_s^{(i)} \}_{i=1}^{N}
\right).
\end{equation}

We set $\alpha_{\mathrm{cp}} = 0.05$, corresponding to highly confident cases in the calibration data. During inference, if $\mathrm{cp}_s(o) \le q_{\mathrm{cp}}$, we consider the prediction sufficiently confident and terminate without further user queries.





\subsubsection{Question Generation via LLM}

For objects selected based on uncertainty, we generate user queries and interpret responses using an LLM. This enables flexible and context-aware natural language interaction.

\paragraph{Query Generation}

For each selected object $o$, the LLM generates a natural-language question to identify its owner. The goal is to produce concise, intuitive queries that users can easily answer.

The LLM takes the following inputs:
\begin{itemize}
    \item object information (\textit{e.g.}, ID, class, location)
    \item candidate user set $\mathcal{U}$
    \item current ownership scores $\mathbf{s}(o)$
\end{itemize}

The ownership scores are provided as multi-label outputs and are not constrained to sum to one. Based on these scores, the LLM focuses on the top-ranked candidates and generates a question that efficiently distinguishes between them, thereby reducing user cognitive load.

The prompt used for query generation is shown in Prompt~\ref{prompt:prompt_question}.



\begin{prompt}{Ownership Question Generation Prompt}{prompt_question}
1: You are a dialogue module for a home-service robot.\\
2: Based on the given information, generate exactly ONE natural English question\\
3: that the robot will ask the user to identify the owner of an object.\\
4:\\
5: The robot needs to ask the user who owns the following object.\\
6:\\
7: \#\#\# Object Information \\
8: [Object ID]\\
9: [\textcolor{teal}{\textbf{OBJECT\_ID}}]\\
10:\\
11: [Object class]\\
12: [\textcolor{teal}{\textbf{OBJECT\_CLASS}}]\\
13:\\
14: [Position information]\\
15: [\textcolor{teal}{\textbf{POSITION}}]\\
16:\\
17: [Ownership candidates]\\
18: [\textcolor{teal}{\textbf{OWNERS}}]\\
19:\\
20: [Ownership probability distribution $P_{\text{final}}$ (if available)]\\
21: [\textcolor{teal}{\textbf{P\_FINAL}}]\\
22:\\
23: --- Requirements for the generated question --- \\
24: - Produce ONLY one single English question sentence.\\
25: - Make it sound natural and conversational, as if a home robot is asking the user directly.\\
26: - If possible, include light contextual details such as the object's class name,\\
27: \ \ so the user can easily understand which object is being referred to.\\
28: - $P_{\text{final}}$ represents estimated multi-label ownership probabilities for each candidate owner\\
29: \ \ (the values do not necessarily sum to 1). Use this distribution to design a question that is\\
30: \ \ easy for the user to answer, for example by focusing on the 1--2 most likely owners as explicit\\
31: \ \ options and treating the others as `someone else' if appropriate.\\
32: - You may mention the ownership candidates ([\textcolor{teal}{\textbf{OWNERS}}]) or `someone else'.\\
33: - The output must be ONLY the question sentence.\\
34: \ \ Do NOT include explanations, bullet points, JSON, quotes, or additional text.
\tcblower
\textcolor{teal}{\textbf{OBJECT\_ID}}: Identifier of the target object\\
\textcolor{teal}{\textbf{OBJECT\_CLASS}}: Class label of the target object\\
\textcolor{teal}{\textbf{POSITION}}: Spatial position of the target object\\
\textcolor{teal}{\textbf{OWNERS}}: Set of candidate owners\\
\textcolor{teal}{\textbf{P\_FINAL}}: Estimated ownership scores

\end{prompt}



\paragraph{User Response Interpretation and Ownership Extraction}

Users respond to generated queries in free-form natural language. We interpret these responses using an LLM to extract ownership information.

The LLM takes the following inputs:
\begin{itemize}
    \item the generated question
    \item the user's response
    \item the candidate user set $\mathcal{U}$
\end{itemize}

Given these inputs, the LLM determines, for each candidate user $u \in \mathcal{U}$, whether the user is an owner of the object. The output is represented as a binary ownership vector:
\begin{equation}
\mathbf{b}(o)
=
\{ b_u(o) \mid u \in \mathcal{U},\; b_u(o) \in \{0,1\} \}.
\end{equation}

Here, $b_u(o) = 1$ indicates that user $u$ is identified as an owner based on the user's response. Multiple users can be assigned positive labels, allowing the representation of shared ownership.

The prompt used for response interpretation is shown in Prompt~\ref{prompt:prompt_recognition}.


\begin{prompt}{Question-Based Ownership Determination Prompt}{prompt_recognition}
1: You are an excellent household robot.\\
2: From the dialogue between the robot and the user, you must decide,\\
3: for each candidate person, whether they are an owner of the object (True) or not (False).\\
4: Multiple people can be owners (shared / joint ownership is allowed).\\
5:\\
6: Below is the robot's question to the user and the user's answer.\\
7:\\
8: [Robot's question]\\
9: [\textcolor{teal}{\textbf{QUESTION}}]\\
10:\\
11: [User's answer]\\
12: [\textcolor{teal}{\textbf{USER\_ANSWER}}]\\
13:\\
14: [Ownership candidates]\\
15: [\textcolor{teal}{\textbf{OWNERS}}]\\
16:\\
17: --- Decision rules --- \\
18: - For each candidate person, return true if they can reasonably be considered an owner of the object,\\
19: \ \ and false if they should not be considered an owner.\\
20: - It is allowed that more than one person is true (shared ownership is possible).\\
21: - If the user mentions someone who is not in the candidate list, you should normally return false\\
22: \ \ for all candidates, unless the user also clearly indicates one of the candidates as an owner.\\
23: - Even if the user's answer is vague, try to make a reasonable true/false decision for each candidate.\\
24: - If it is truly impossible to decide, you may return false for all candidates.\\
25:\\
26: --- Output format --- \\
27: Return ONLY the following JSON. Do NOT include any extra text, explanation, or natural language.\\
28: \{\{\\
29: \ \ "ownership\_boolean": \{\\
30: \ \ \ \ "CANDIDATE\_1": true/false,\\
31: \ \ \ \ "CANDIDATE\_2": true/false,\\
32: \ \ \ \ \dots\\
33: \ \ \}\\
34: \}\}
\tcblower
\textcolor{teal}{\textbf{QUESTION}}: Question presented to the user by the robot\\
\textcolor{teal}{\textbf{USER\_ANSWER}}: User's response to the question\\
\textcolor{teal}{\textbf{OWNERS}}: Set of candidate owner names
\end{prompt}



\section{Experiment I: Evaluation of Ownership Inference Performance}\label{sec:experiment1}

The objective of this experiment is to evaluate whether the proposed method can accurately estimate object ownership in a simulated everyday environment, while reducing the number of user queries required for inference.


\subsection{Experimental Conditions}\label{subsec:conditions}

We construct a simulated everyday environment in which multiple users interact with objects through usage, movement, and sharing. The environment contains a variety of objects whose ownership must be inferred based on contextual information rather than simple spatial proximity.

To evaluate context-dependent ownership inference, we include objects and interactions with the following characteristics:
\begin{itemize}
    \item \textbf{Context-dependent objects}: Objects strongly associated with user background (\textit{e.g.}, a PC associated with a working father, or toys preferred by children), allowing us to evaluate the impact of user roles and occupations.
    \item \textbf{Usage-based interactions}: Actions such as carrying cups or reading books, enabling evaluation of how usage frequency and recency contribute to ownership estimation.
\end{itemize}

\begin{figure}[tb]
    \centering
    \includegraphics[width=0.6\linewidth]{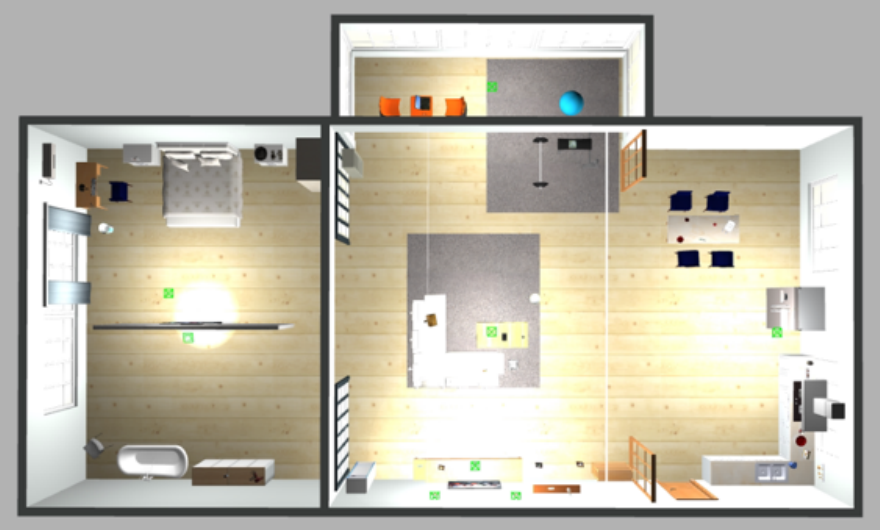}
    \caption{Overview of the experimental environment.}
    \label{fig:environment}
\end{figure}

The experiments are conducted in a Gazebo-based simulation environment using the \textit{aws-robomaker-small-house-world}~\footnote{https://github.com/aws-robotics/aws-robomaker-small-house-world}. An overview of the environment is shown in Figure~\ref{fig:environment}.

We use the Household Object Movements from Everyday Routines (HOMER) dataset~\cite{HOMER}, which provides time-series records of object usage and movement associated with daily human activities. We use 7 days of data (6:00--23:00), comprising 3,425 interaction events across 34 objects.

Ground-truth ownership is manually assigned for each object, including both single-user ownership and shared ownership cases.

\begin{figure}[tb]
    \centering
    \includegraphics[width=\linewidth]{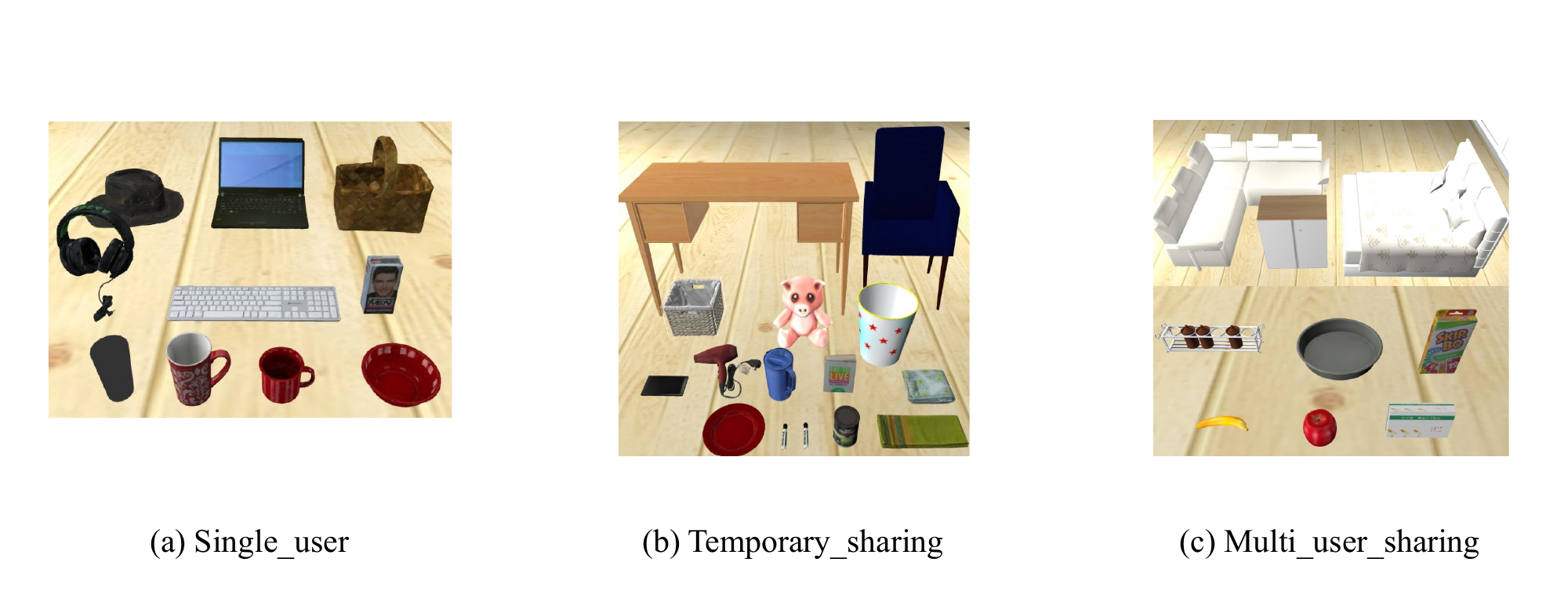}
    \caption{List of objects used in this experiment}
    \label{fig:dataset}
\end{figure}

To evaluate different ownership scenarios, we design the dataset to include the following conditions:
\begin{itemize}
    \item \textbf{Single\_user}: Objects used exclusively by a single owner (Figure~\ref{fig:dataset}(a))
    \item \textbf{Temporary\_sharing}: Objects temporarily used by non-owners (Figure~\ref{fig:dataset}(b))
    \item \textbf{Multi\_user\_sharing}: Objects used by multiple users with similar frequency (Figure~\ref{fig:dataset}(c))
\end{itemize}

We consider a setting with three users:
\begin{itemize}
    \item Bob (father, office worker)
    \item Mary (mother, homemaker)
    \item Tom (son, elementary school student)
\end{itemize}

We use GPT-4o (gpt-4o-2024-11-20) as the LLM. The prompt is fixed across all trials, and the temperature is set to 0.2 to ensure stable and reproducible outputs.

The main parameters of the proposed method are set as follows. 
For spatial context extraction, we set $\gamma = 1.0$, corresponding to the isotropic Euclidean distance in 3D, and $\sigma = 0.5$ to control the spatial decay. The numbers of neighboring and similar objects are set to $K_{\mathrm{near}} = 5$ and $K_{\mathrm{sim}} = 5$, respectively.

For conformal prediction, we set $\alpha = 0.2$, ensuring that the true owner is included in the prediction set with probability at least $80\%$. 
The stopping threshold is defined using $\alpha_{\mathrm{cp}} = 0.05$, resulting in a conservative criterion that terminates interaction only when high confidence is achieved.

For shared ownership detection, the thresholds are set to $\varepsilon_{\mathrm{min}} = 0.80$, $\varepsilon_{\mathrm{in}} = 0.08$, and $\varepsilon_{\mathrm{out}} = 0.20$, ensuring both high confidence within shared users and clear separation from non-shared users. 
The maximum number of queries $Q_{\max}$ was set to the number of objects.

All experiments are conducted on a machine equipped with an Intel Xeon Gold 6548Y+ (32 cores), 125 GB RAM, and an NVIDIA RTX 6000 Ada GPU (48 GB), running Ubuntu 22.04.5 LTS with a ROS Noetic-based Docker environment~\cite{el_hafi_software_2022}.



\subsection{Dataset Preparation}

To construct a dataset that explicitly includes user–object interactions and ownership information, we extend the HOMER dataset~\cite{HOMER}, which provides time-series records of object movements in household environments.

Since HOMER does not include ownership labels or user-specific interactions, we generate a usage history dataset through a structured preprocessing pipeline (see Appendix~\ref{app:dataset_generation} for details). First, we sample $N=7$ days of activity schedules and expand them into sequences of interaction events. Each event is represented as a tuple
\[
e_i = (d_i, t_i, u_i, a_i, o_i),
\]
where $d_i$, $t_i$, $u_i$, $a_i$, and $o_i$ denote the date, time, user, action, and object, respectively. These events are converted into natural language descriptions and organized into time-ordered usage histories for each object.

We then assign ground-truth ownership labels $O(o)$ to each object, including both single-user and shared ownership. To simulate realistic usage patterns, we consider three scenarios:
\begin{itemize}
    \item \textbf{Single-user}: only the owner uses the object
    \item \textbf{Temporary sharing}: non-owners occasionally use the object
    \item \textbf{Multi-user sharing}: multiple users use the object with similar frequency
\end{itemize}

Interaction events are grouped into usage sessions, and the resulting dataset is split into training and evaluation sets based on temporal order: the first 3 days are used for training, and the remaining data for evaluation.

This process yields a dataset that preserves temporal dynamics and explicitly models user–object relationships, enabling realistic evaluation of ownership inference in everyday environments.

\subsection{Comparison Methods}

To evaluate the proposed method, we compare it with the following baselines. All methods are evaluated using the same usage history data and user settings. A summary is provided in Table~\ref{tab:comparison_methods}.

\begin{enumerate}

\item \textbf{ActOwL~\cite{ActOwL}}:
An interactive ownership estimation method based on information gain (IG) that actively queries users to acquire ownership information. This baseline allows us to evaluate the effectiveness of our uncertainty-driven questioning and contextual integration compared to an existing interactive approach.

\item \textbf{NLMap~\cite{NLMap} + LLM}:
A method that estimates ownership using only object-centric information from NLMap (\textit{e.g.}, location and visual features) as input to the LLM. This baseline evaluates the contribution of incorporating contextual information such as user background and usage history.

\item \textbf{Last-User (Tan~\textit{et~al.}~\cite{tan2019s}-style)}:
A heuristic baseline that assigns ownership to the most recent user of the object. This method evaluates the limitations of relying solely on recency-based interaction signals.

\item \textbf{Frequency-Based (Wu~\textit{et~al.}~\cite{wu2020item}-style)}:
A heuristic baseline that assigns ownership to the user with the highest interaction frequency. This baseline evaluates the effectiveness of context-aware reasoning compared to simple aggregation-based methods.

\end{enumerate}





\begin{table}[t]
\centering
\caption{Comparison of Methods and Their Capabilities}
\label{tab:comparison_methods}
\begin{tabular}{lccc}
\hline
Method &
\makecell{Active\\Questioning} &
\makecell{User Background\\Information} &
\makecell{Usage\\History} \\
\hline
ActOwL~\cite{ActOwL} & $\checkmark$ (IG)  & --         & -- \\
NLMap~\cite{NLMap} + LLM   & --          & --         & -- \\
Last-User~\cite{tan2019s}  & --          & --         & $\checkmark$ \\
Frequency-Based~\cite{wu2020item} & --   & --         & $\checkmark$ \\
COIN (ours)            & $\checkmark$ (CP)  & $\checkmark$ & $\checkmark$ \\
\hline
\end{tabular}
\end{table}

\subsection{Evaluation Metrics}

We evaluate the proposed method from two perspectives: (i) ownership inference accuracy and (ii) interaction efficiency.
For ownership inference, we use standard multi-label evaluation metrics, including Subset Accuracy, Mean Jaccard, and Micro-averaged Precision, Recall, and F1 score. 
These metrics compare the predicted ownership set with the ground-truth ownership set for each object, capturing both exact-match and partial-match performance.
To evaluate interaction efficiency, we measure the number of questions needed to determine an object's ownership. 
A smaller number of questions indicates lower user burden and more efficient information acquisition.
Since different methods produce outputs in different formats, all outputs are converted into a unified ownership set representation before evaluation. Detailed definitions of each metric are provided in Appendix~\ref{app:metrics}.

\subsection{Qualitative Results}

We qualitatively analyze the behavior of the proposed method to understand how ownership estimates evolve through interaction and how the generated queries contribute to reducing uncertainty.

\subsubsection{Evolution of Ownership Scores}

\begin{figure}[tb]
    \centering
    \includegraphics[width=1.0\linewidth]{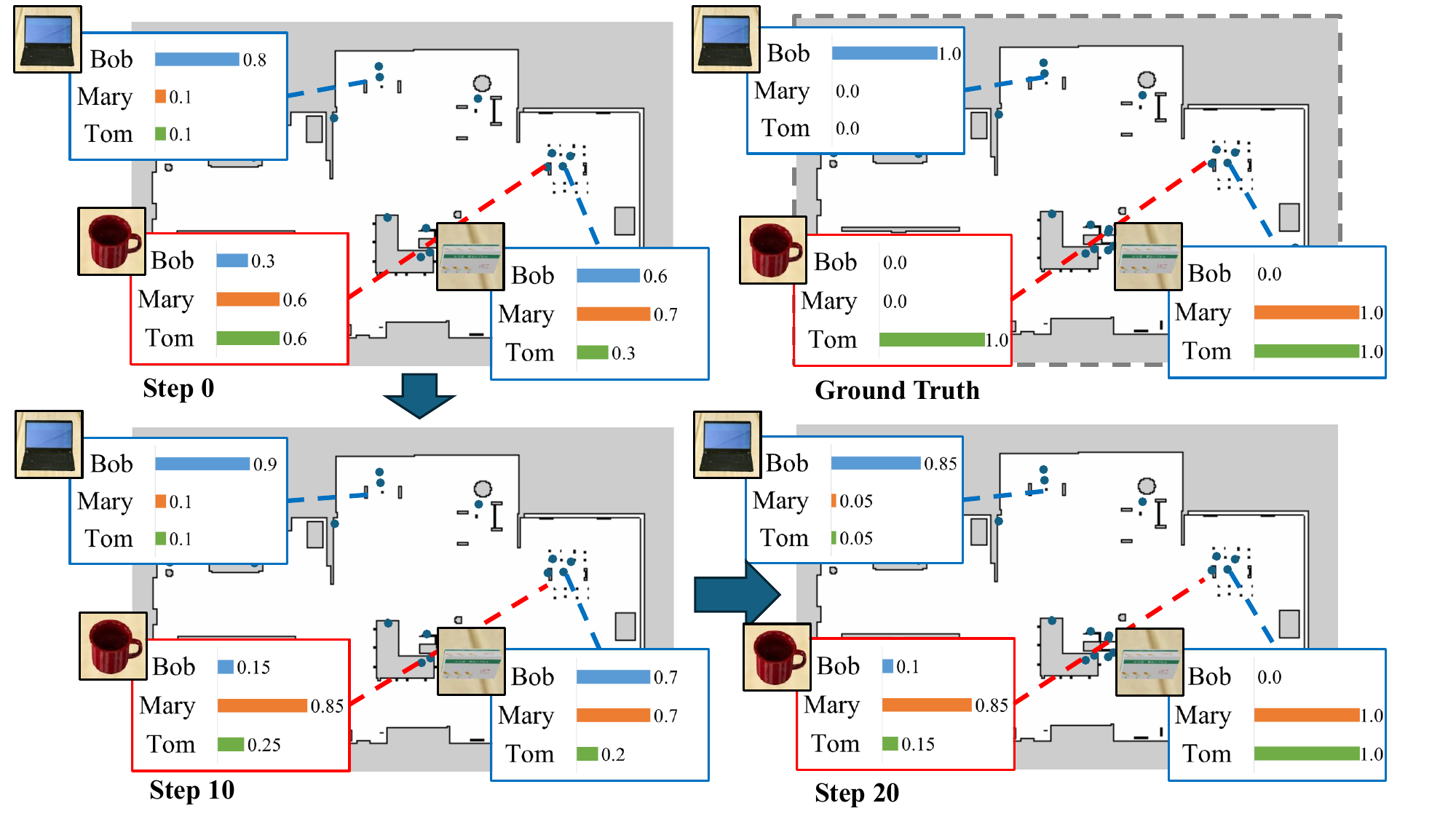}
    \caption{
    Temporal evolution of ownership scores produced by the LLM across query steps.
    Scores progressively concentrate on the correct owner as interaction proceeds.
    Blue indicates correctly inferred objects, while red indicates incorrect ones.
    }
    \label{fig:llm_output}
\end{figure}

Figure~\ref{fig:llm_output} illustrates the temporal evolution of ownership scores for representative objects at Step 0, Step 10, and Step 20.

At Step 0, the LLM produces diffuse predictions, assigning non-negligible scores to multiple candidates. For example, the \textit{cup} is ambiguous between Mary and Tom, while the \textit{laptop} already shows a strong bias toward Bob.

As interaction proceeds, the scores become increasingly concentrated on the correct owner. By Step 10, the correct candidates dominate, although residual uncertainty remains. By Step 20, the scores converge close to 1.0 for the correct owners, indicating that uncertainty is effectively reduced through iterative updates and selective questioning.

However, failure cases are also observed. For instance, the \textit{cup} is incorrectly assigned to Mary, even though Tom is the ground-truth owner. This occurs because contextual dependencies across objects can reinforce incorrect hypotheses, leading to globally consistent but locally incorrect predictions.

These results demonstrate that the proposed method progressively refines ownership estimates by integrating contextual information and user feedback, while also highlighting the risk of error propagation through contextual reasoning.






\subsubsection{Generated Questions and Response Interpretation}

\begin{figure}[tb]
    \centering
    \includegraphics[width=1.0\linewidth]{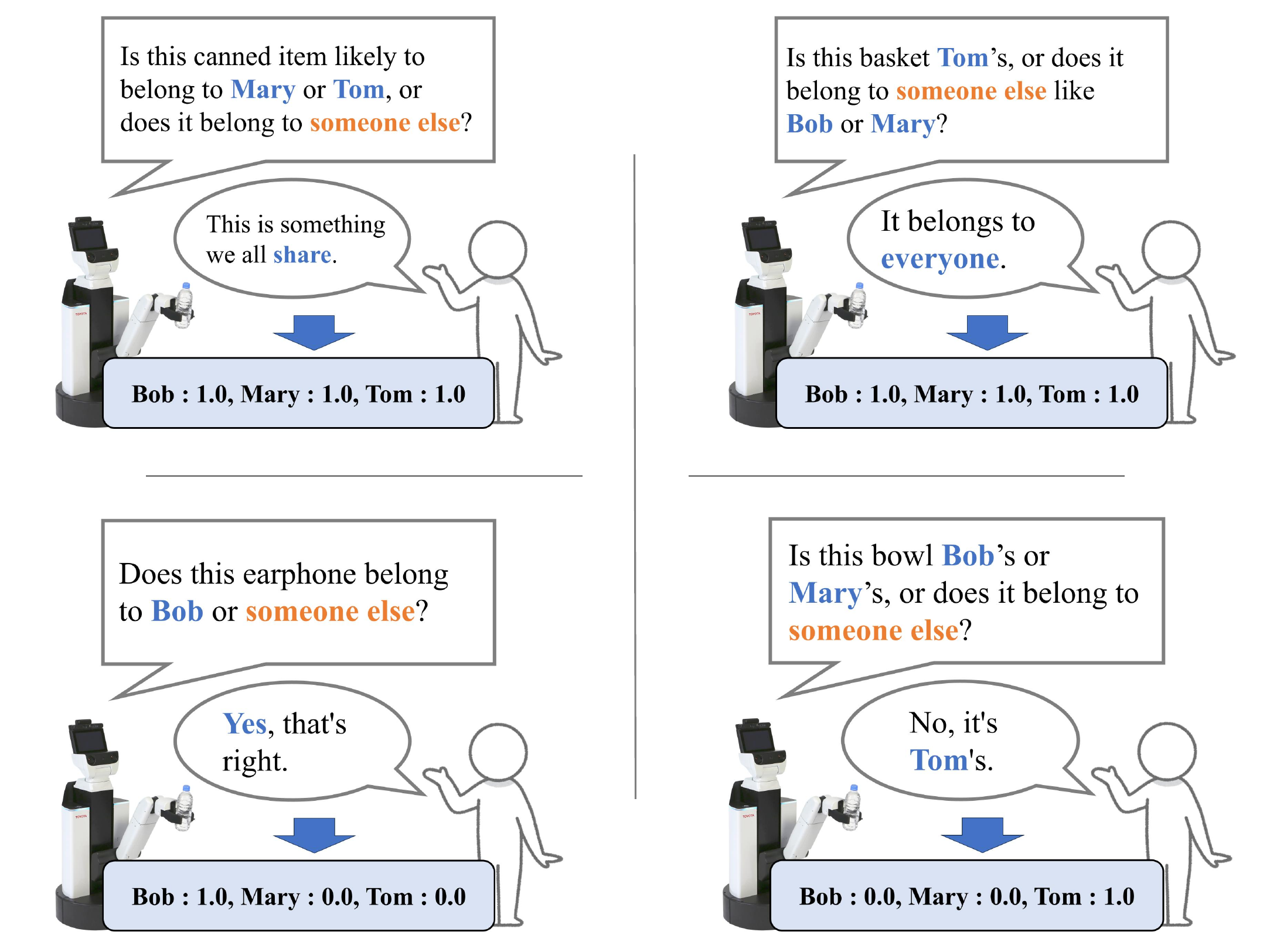}
    \caption{
    Example of LLM-based question generation and response interpretation.
    The system generates a natural language query based on ownership uncertainty, and the user's response is subsequently interpreted to update ownership predictions.
    }
    \label{fig:question_generation}
\end{figure}

Figure~\ref{fig:question_generation} shows examples of generated questions and corresponding interpretations.
The proposed method generates natural and concise questions by explicitly presenting the most probable candidates while grouping the remaining ones into a generic alternative (\textit{e.g.}, `someone else'). This design limits the number of choices and reduces user cognitive load.

In practice, the questions consistently include one or two high-confidence candidates along with the object name, enabling users to respond intuitively. The generated expressions are natural and suitable for human–robot interaction in everyday environments.

These results indicate that the method effectively translates probabilistic ownership estimates into user-friendly queries, facilitating efficient information acquisition.


Additional analysis of query target selection is provided in Appendix~\ref{app:query_analysis}.



\subsection{Quantitative Evaluation Results}

We quantitatively evaluate the proposed method in terms of ownership estimation accuracy, interaction efficiency, and computational cost. The results demonstrate the effectiveness of integrating contextual information and uncertainty-driven interaction.

\subsubsection{Overall Performance}

Table~\ref{tab:result} reports the mean and standard deviation over 10 trials.
For the final step performance, statistical significance is evaluated using the Kruskal–Wallis test, followed by Mann–Whitney U tests with Holm correction.
The proposed method achieves the best performance across all major metrics, including Subset Accuracy, Mean Jaccard, Micro Recall, and Micro F1.
In particular, the improvement in Subset Accuracy indicates that the method accurately captures ownership as a consistent set.

\subsubsection{Comparison with Baselines}

ActOwL achieves high Micro Precision but suffers from lower recall, indicating incomplete ownership estimation.
Usage-based methods show moderate performance, suggesting that usage history provides a strong signal but is insufficient in complex scenarios such as temporary sharing.
NLMap+LLM performs poorly, highlighting the limitation of object-centric features without contextual information.

\subsubsection{Effect of Interactive Questioning}

\begin{figure}[tb]
    \centering
    \includegraphics[width=0.8\linewidth]{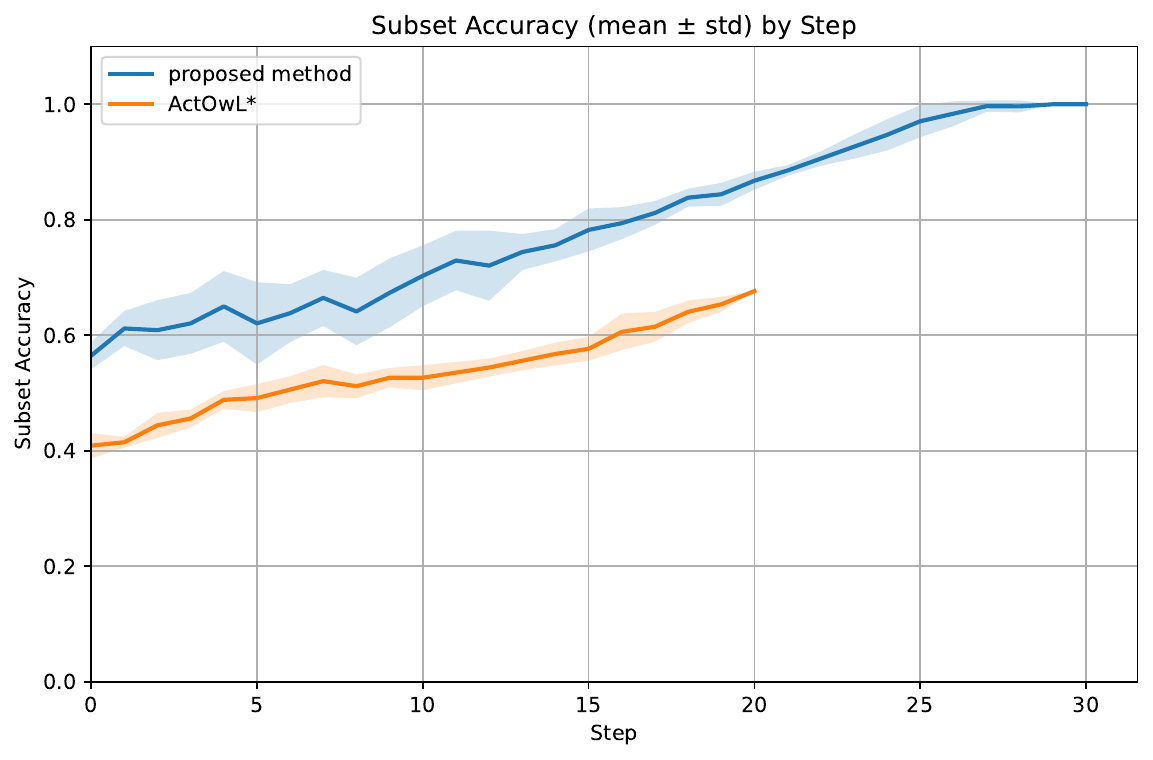}
    \caption{
    Subset Accuracy across query steps for active questioning methods (mean $\pm$ standard deviation over 10 trials). 
    Statistically significant differences compared to the proposed method are indicated by $^{*}$ ($p<0.05$).
    }
    \label{fig:result_vs_ActOwL}
\end{figure}

Figure~\ref{fig:result_vs_ActOwL} shows the evolution of Subset Accuracy across query steps.
Both methods improve with interaction, but the proposed method achieves higher final accuracy and more stable convergence.
Kaplan--Meier analysis further shows that the proposed method reaches the target accuracy more reliably.

\subsubsection{Efficiency Analysis}

Table~\ref{tab:question_inference_result} summarizes the number of queries, inference time, and token usage.
The proposed method requires more queries but achieves significantly higher accuracy, indicating a favorable trade-off.
It also reduces inference time compared to ActOwL while increasing token usage through richer contextual reasoning.

\begin{table}[tb]
  \centering
  \caption{Mean and standard deviation over 10 trials for each evaluation metric (mean $\pm$ standard deviation)}
  \label{tab:result}
  \resizebox{\linewidth}{!}{%
  \begin{tabular}{lccccc}
    \hline
    Method &
    Subset Accuracy $\uparrow$ &
    Mean Jaccard $\uparrow$ &
    Micro Precision $\uparrow$ &
    Micro Recall $\uparrow$ &
    Micro F1 $\uparrow$
    \\
    \hline
    ActOwL~\cite{ActOwL}
    & \underline{$0.674 \pm 0.009$}\sym{*}
    & $0.674 \pm 0.009$\sym{*}
    & \underline{$\mathbf{1.000 \pm 0.000}$}
    & $0.529 \pm 0.023$\sym{*}
    & $0.691 \pm 0.020$\sym{*}
    \\
    
    NLMap~\cite{NLMap}+LLM
    & $0.182 \pm 0.043$\sym{*}
    & $0.367 \pm 0.043$\sym{*}
    & $0.539 \pm 0.033$\sym{*}
    & $0.444 \pm 0.039$\sym{*}
    & $0.486 \pm 0.034$\sym{*}
    \\
    
    Tan~\textit{et~al.}~\cite{tan2019s}
    & $0.588 \pm 0.000$\sym{*}
    & $0.721 \pm 0.000$\sym{*}
    & $0.941 \pm 0.000$\sym{*}
    & $0.582 \pm 0.000$\sym{*}
    & $0.719 \pm 0.000$\sym{*}
    \\

    Wu~\textit{et~al.}~\cite{wu2020item}
    & $0.647 \pm 0.000$\sym{*}
    & \underline{$0.779 \pm 0.000$}\sym{*}
    & \underline{$\mathbf{1.000 \pm 0.000}$}
    & \underline{$0.618 \pm 0.000$}\sym{*}
    & \underline{$0.764 \pm 0.000$}\sym{*}
    \\
    
    COIN (ours)
    & \underline{$\mathbf{0.988 \pm 0.028}$}
    & \underline{$\mathbf{0.991 \pm 0.023}$}
    & \underline{$0.996 \pm 0.012$}
    & \underline{$\mathbf{0.993 \pm 0.018}$}
    & \underline{$\mathbf{0.994 \pm 0.014}$}
    \\
    \hline
  \end{tabular}
}
\vspace{1mm}
\begin{flushleft}
\footnotesize
$^{*}p<0.05$．
An asterisk indicates a statistically significant difference compared with the proposed method.
\end{flushleft}
\end{table}

\begin{table}[tb]
  \centering
  \caption{Number of questions until convergence, inference time, and token usage of active question generation methods (mean $\pm$ standard deviation)}
  \label{tab:question_inference_result}
  \begin{tabular}{lccc}
    \hline
    Method & 
    \makecell{Number of\\Questions $\downarrow$} & 
    \makecell{Inference Time (s) $\downarrow$} & 
    \makecell{Number of Tokens $\downarrow$} \\
    \hline
    ActOwL~\cite{ActOwL}
    & $\mathbf{19.7 \pm 0.949}$
    & $1530.144 \pm 125.125$
    & $\mathbf{326{,}900 \pm 3{,}479}$ \\
    
    COIN (ours)
    & $28.4 \pm 1.578$ 
    & $\mathbf{625.831 \pm 27.996}$
    & $511{,}706 \pm 6{,}862$ \\
    \hline
  \end{tabular}
\end{table}

\subsection{Results by Object Usage Type}

Tables~\ref{tab:single_user_result}, \ref{tab:temporary_sharing_result}, and \ref{tab:multi_user_sharing_result} report the results for different object usage settings: single-user, temporary sharing, and multi-user sharing.

Overall, the proposed method achieves consistently high performance across all settings, demonstrating robustness to different ownership conditions. In contrast, baseline methods exhibit significant performance variations across usage scenarios.

\subsubsection{Single-user setting}
In the single-user setting, rule-based methods (Tan~\textit{et~al.} and Wu~\textit{et~al.}) achieve perfect performance, as their assumptions align with the environment. 
The proposed method also achieves near-perfect performance, confirming that it does not sacrifice accuracy even under simplified conditions. 
In contrast, ActOwL and NLMap+LLM perform worse due to errors in early-stage decisions and a lack of contextual reasoning.

\subsubsection{Temporary sharing setting}
In the temporary sharing setting, the proposed method significantly outperforms all baselines across all metrics. 
This result highlights its ability to distinguish between temporary use and true ownership by integrating contextual information and user interactions. 
In contrast, rule-based methods degrade in performance because they assume consistency between usage and ownership, and ActOwL shows unstable performance due to the absence of contextual reasoning.

\subsubsection{Multi-user sharing setting}
In the multi-user sharing setting, the proposed method achieves perfect performance across all metrics, demonstrating its capability to accurately model shared ownership. 
ActOwL achieves relatively high precision but suffers from low recall, indicating difficulty in capturing all true owners. Rule-based methods fail to represent shared ownership, while NLMap+LLM struggles to produce stable multi-label predictions.

These results confirm that the proposed method generalizes across different ownership structures and is particularly effective in complex scenarios involving sharing and temporal usage.

\subsubsection{Effect of Interaction Across Usage Types}

\begin{figure}[tb]
    \centering
    \includegraphics[width=0.8\linewidth]{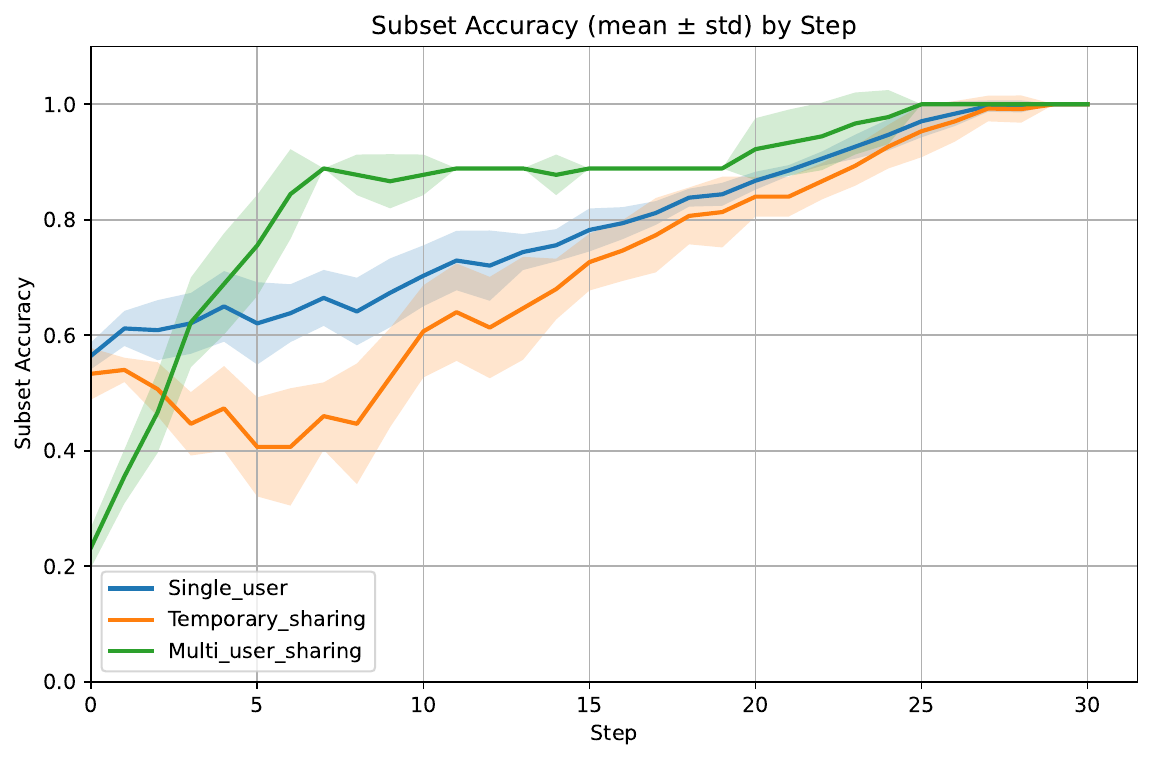}
    \caption{
    Subset Accuracy across query steps for different object usage types (mean $\pm$ standard deviation over 10 trials).
    Single\_user denotes objects used exclusively by their owners, 
    Temporary\_sharing denotes objects with temporary borrowing, and 
    Multi\_user\_sharing denotes objects used by multiple users with comparable frequency.
    }
    \label{fig:result_object_setting}
\end{figure}

Figure~\ref{fig:result_object_setting} shows the evolution of Subset Accuracy across query steps for each usage type.
In all settings, accuracy improves as interaction proceeds, demonstrating the effectiveness of question-driven refinement.

In the single-user setting, high accuracy is achieved with only a few queries due to strong initial cues from context. 
In contrast, the temporary sharing setting starts with lower accuracy but steadily improves, indicating that interaction helps distinguish ownership from transient usage.

The multi-user sharing setting exhibits rapid improvement after initial steps, suggesting that the proposed method effectively captures shared ownership through early interaction.

We also observe temporary drops in accuracy in the early stages for single-user and temporary sharing settings. 
This is due to the prioritization of uncertain objects, which temporarily increases ambiguity before sufficient information is acquired. Despite this, the method converges to highly accurate predictions in later steps.






\begin{table}[tb]
\centering
\caption{Mean and standard deviation over 10 trials for each evaluation metric on objects categorized as Single\_user (mean $\pm$ standard deviation)．
}
\label{tab:single_user_result}
\resizebox{\linewidth}{!}{%
\begin{tabular}{lccccc}
\hline
Method &
Subset Accuracy $\uparrow$ &
Mean Jaccard $\uparrow$ &
Micro Precision $\uparrow$ &
Micro Recall $\uparrow$ &
Micro F1 $\uparrow$ \\
\hline
ActOwL~\cite{ActOwL}
& $0.800 \pm 0.000$\sym{*}
& $0.800 \pm 0.000$\sym{*}
& $0.800 \pm 0.000$\sym{*}
& $0.800 \pm 0.000$\sym{*}
& $0.800 \pm 0.000$\sym{*}
\\
NLMap~\cite{NLMap}+LLM
& $0.250 \pm 0.053$\sym{*}
& $0.313 \pm 0.066$\sym{*}
& $0.328 \pm 0.067$\sym{*}
& $0.390 \pm 0.088$\sym{*}
& $0.356 \pm 0.075$\sym{*}
\\
Tan~\textit{et~al.}~\cite{tan2019s}
& \underline{$\mathbf{1.000 \pm 0.000}$}
& \underline{$\mathbf{1.000 \pm 0.000}$}
& \underline{$\mathbf{1.000 \pm 0.000}$}
& \underline{$\mathbf{1.000 \pm 0.000}$}
& \underline{$\mathbf{1.000 \pm 0.000}$}
\\
Wu~\textit{et~al.}~\cite{wu2020item}
& \underline{$\mathbf{1.000 \pm 0.000}$}
& \underline{$\mathbf{1.000 \pm 0.000}$}
& \underline{$\mathbf{1.000 \pm 0.000}$}
& \underline{$\mathbf{1.000 \pm 0.000}$}
& \underline{$\mathbf{1.000 \pm 0.000}$}
\\
COIN (ours)
& \underline{$0.980 \pm 0.063$}
& \underline{$0.980 \pm 0.063$}
& \underline{$0.980 \pm 0.063$}
& \underline{$0.980 \pm 0.063$}
& \underline{$0.980 \pm 0.063$}
\\
\hline
\end{tabular}
}
\vspace{1mm}
\begin{flushleft}
\footnotesize
$^{*}p<0.05$．
An asterisk indicates a statistically significant difference compared with the proposed method.
\end{flushleft}
\end{table}

\begin{table}[tb]
\centering
\caption{Mean and standard deviation over 10 trials for each evaluation metric on objects categorized as Temporary\_sharing (mean $\pm$ standard deviation)．
}
\label{tab:temporary_sharing_result}
\resizebox{\linewidth}{!}{%
\begin{tabular}{lccccc}
\hline
Method &
Subset Accuracy $\uparrow$ &
Mean Jaccard $\uparrow$ &
Micro Precision $\uparrow$ &
Micro Recall $\uparrow$ &
Micro F1 $\uparrow$ \\
\hline
ActOwL~\cite{ActOwL}
& $0.667 \pm 1.170$\sym{*}
& $0.667 \pm 1.170$\sym{*}
& $0.667 \pm 1.170$\sym{*}
& \underline{$0.769 \pm 0.000$}\sym{*}
& $0.714 \pm 0.000$\sym{*}
\\
NLMap~\cite{NLMap}+LLM
& $0.133 \pm 0.089$\sym{*}
& $0.304 \pm 0.082$\sym{*}
& $0.420 \pm 0.055$\sym{*}
& $0.422 \pm 0.077$\sym{*}
& $0.420 \pm 0.063$\sym{*}
\\
Tan~\textit{et~al.}~\cite{tan2019s}
& $0.600 \pm 0.000$\sym{*}
& $0.700 \pm 0.000$\sym{*}
& \underline{$0.867 \pm 0.000$}\sym{*}
& $0.591 \pm 0.000$\sym{*}
& $0.703 \pm 0.000$\sym{*}
\\
Wu~\textit{et~al.}~\cite{wu2020item}
& \underline{$0.733 \pm 0.000$}\sym{*}
& \underline{$0.833 \pm 0.000$}\sym{*}
& \underline{$\mathbf{1.000 \pm 0.000}$}
& $0.682 \pm 0.000$\sym{*}
& \underline{$0.811 \pm 0.000$}\sym{*}
\\
COIN (ours)
& \underline{$\mathbf{0.987 \pm 0.028}$}
& \underline{$\mathbf{0.993 \pm 0.014}$}
& \underline{$\mathbf{1.000 \pm 0.000}$}
& \underline{$\mathbf{0.991 \pm 0.019}$}
& \underline{$\mathbf{0.995 \pm 0.010}$}
\\
\hline
\end{tabular}
}
\vspace{1mm}
\begin{flushleft}
\footnotesize
$^{*}p<0.05$．
An asterisk indicates a statistically significant difference compared with the proposed method.
\end{flushleft}
\end{table}

\begin{table}[tb]
\centering
\caption{Mean and standard deviation over 10 trials for each evaluation metric on objects categorized as Multi\_user\_sharing (mean $\pm$ standard deviation)．
}
\label{tab:multi_user_sharing_result}
\resizebox{\linewidth}{!}{%
\begin{tabular}{lccccc}
\hline
Method &
Subset Accuracy $\uparrow$ &
Mean Jaccard $\uparrow$ &
Micro Precision $\uparrow$ &
Micro Recall $\uparrow$ &
Micro F1 $\uparrow$ \\
\hline
ActOwL~\cite{ActOwL}
& \underline{$0.878 \pm 0.035$}\sym{*}
& \underline{$0.878 \pm 0.035$}\sym{*}
& \underline{$\mathbf{1.000 \pm 0.000}$}
& \underline{$0.560 \pm 0.126$}\sym{*}
& \underline{$0.708 \pm 0.132$}\sym{*}
\\
NLMap~\cite{NLMap}+LLM
& $0.189 \pm 0.075$\sym{*}
& $0.531 \pm 0.030$\sym{*}
& \underline{$0.992 \pm 0.026$}
& $0.487 \pm 0.027$\sym{*}
& $0.653 \pm 0.026$\sym{*}
\\
Tan~\textit{et~al.}~\cite{tan2019s}
& $0.111 \pm 0.000$\sym{*}
& $0.444 \pm 0.000$\sym{*}
& \underline{$\mathbf{1.000 \pm 0.000}$}
& $0.391 \pm 0.000$\sym{*}
& $0.563 \pm 0.000$\sym{*}
\\
Wu~\textit{et~al.}~\cite{wu2020item}
& $0.111 \pm 0.000$\sym{*}
& $0.444 \pm 0.000$\sym{*}
& \underline{$\mathbf{1.000 \pm 0.000}$}
& $0.391 \pm 0.000$\sym{*}
& $0.563 \pm 0.000$\sym{*}
\\
COIN (ours)
& \underline{$\mathbf{1.000 \pm 0.000}$}
& \underline{$\mathbf{1.000 \pm 0.000}$}
& \underline{$\mathbf{1.000 \pm 0.000}$}
& \underline{$\mathbf{1.000 \pm 0.000}$}
& \underline{$\mathbf{1.000 \pm 0.000}$}
\\
\hline
\end{tabular}
}
\vspace{1mm}
\begin{flushleft}
\footnotesize
$^{*}p<0.05$．
An asterisk indicates a statistically significant difference compared with the proposed method.
\end{flushleft}
\end{table}

\section{Experiment II: Ablation Study}\label{sec:experiment2}

The purpose of this ablation study is to evaluate the contribution of each component in the proposed method by systematically removing them and comparing performance.

\subsection{Ablation Setup}

The experimental environment, dataset, evaluation metrics, and parameter settings are identical to those in Section~\ref{sec:experiment1}. This ensures that any performance differences arise solely from the presence or absence of each component.
The ablation variants are summarized in Table~\ref{tab:ablation_methods}:
\begin{enumerate}
\item \textbf{w/o Active Questioning}: The model performs a single-shot inference without querying the user, relying only on observed and contextual information.
\item \textbf{w/o User Background}: The model excludes user background information and relies on object observations, usage history, and interaction.
\item \textbf{w/o Usage History}: The model excludes usage history and relies on object observations, user background information, and interaction.
\end{enumerate}




\begin{table}[t]
\centering
\caption{Ablation Methods}
\label{tab:ablation_methods}
\begin{tabular}{lccc}
\hline
Method &
\makecell{Active\\Questioning} &
\makecell{User Background\\Information} &
\makecell{Usage\\History} \\
\hline
w/o Active Questioning & -- & $\checkmark$ & $\checkmark$ \\
w/o User Background & $\checkmark$ & -- & $\checkmark$ \\
w/o Usage History & $\checkmark$ & $\checkmark$ & -- \\
COIN (ours) & $\checkmark$ & $\checkmark$ & $\checkmark$ \\
\hline
\end{tabular}
\end{table}

\subsection{Ablation Results}

\begin{table}[tb]
\centering
\caption{Mean and standard deviation over 10 trials for each evaluation metric in the ablation study (mean $\pm$ standard deviation)}
\label{tab:ablation_result}
\resizebox{\linewidth}{!}{%
\begin{tabular}{lcccccc}
\hline
Method &
Subset Accuracy $\uparrow$ &
Mean Jaccard $\uparrow$ &
Micro Precision $\uparrow$ &
Micro Recall $\uparrow$ &
Micro F1 $\uparrow$ &
\makecell{Number of\\Questions $\downarrow$} \\
\hline
w/o Active Questioning
& $0.574 \pm 0.016$\sym{*}
& $0.752 \pm 0.012$\sym{*}
& $0.888 \pm 0.011$\sym{*}
& $0.675 \pm 0.016$\sym{*}
& $0.766 \pm 0.011$\sym{*}
& -- \\

w/o User Background
& \underline{$0.985 \pm 0.016$}
& $0.990 \pm 0.010$
& $0.982 \pm 0.018$
& \underline{$\mathbf{1.000 \pm 0.000}$}
& \underline{$0.991 \pm 0.009$}
& $33.3 \pm 0.483$ \\

w/o Usage History
& \underline{$\mathbf{0.988 \pm 0.028}$}
& \underline{$\mathbf{0.992 \pm 0.017}$}
& \underline{$0.988 \pm 0.027$}
& \underline{$\mathbf{1.000 \pm 0.000}$}
& \underline{$\mathbf{0.994 \pm 0.014}$}
& \underline{$32.5 \pm 1.179$} 
\\

COIN (ours)
& \underline{$\mathbf{0.988 \pm 0.028}$}
& \underline{$0.991 \pm 0.023$}
& \underline{$\mathbf{0.996 \pm 0.012}$}
& \underline{$0.993 \pm 0.018$}
& \underline{$\mathbf{0.994 \pm 0.014}$}
& \underline{$\mathbf{28.4 \pm 1.578}$} 
\\
\hline
\end{tabular}
}
\vspace{1mm}
\begin{flushleft}
\footnotesize
$^{*}p<0.05$．
An asterisk indicates a statistically significant difference compared with the proposed method.
\end{flushleft}
\end{table}

\begin{figure}[tb]
    \centering
    \includegraphics[width=0.8\linewidth]{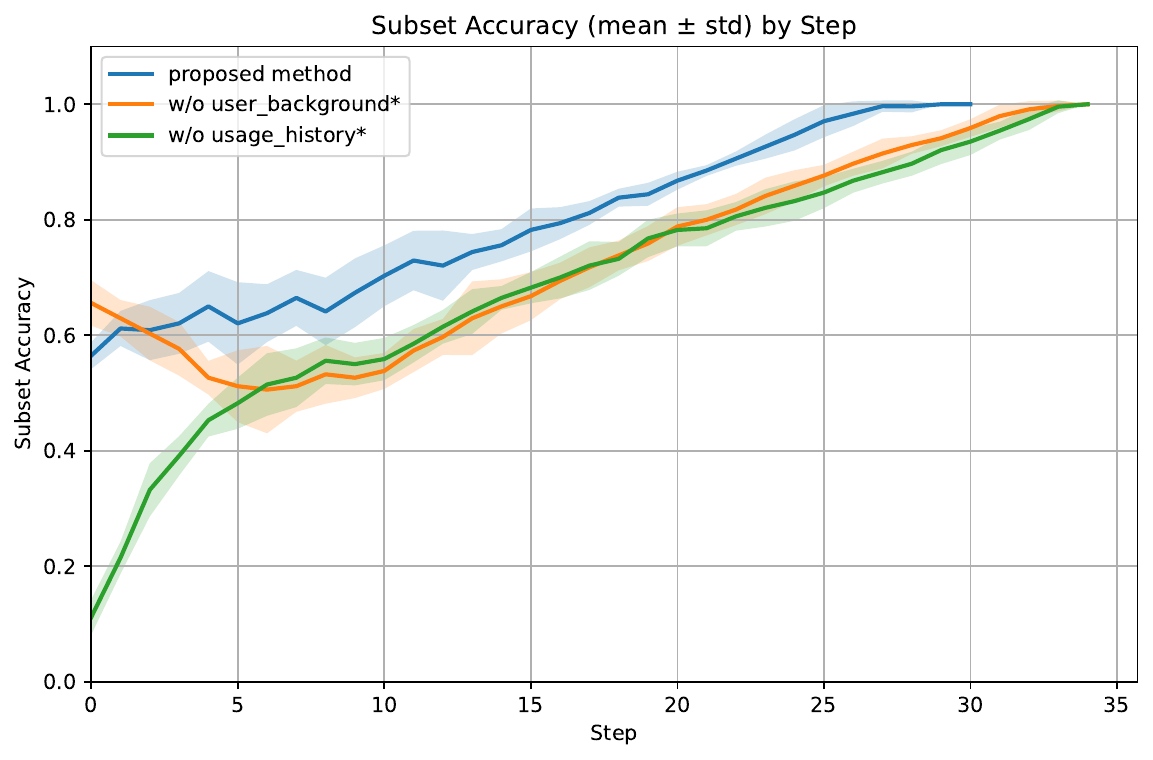}
    \caption{Subset Accuracy at each query step in the ablation study (mean $\pm$ standard deviation over 10 trials). $^{*}p<0.05$ indicates a statistically significant difference compared with the proposed method.}
    \label{fig:result_ablation_step}
\end{figure}

Table~\ref{tab:ablation_result} reports the mean and standard deviation over 10 trials for each ablation setting.
Removing active questioning leads to a substantial drop in performance across all metrics. 
This result indicates that contextual information alone is insufficient for accurate ownership inference and that interactive querying is essential for identifying the correct ownership set.
When user background information is removed, the model maintains relatively high accuracy but requires more queries compared to the full model. 
This suggests that background information helps reduce uncertainty early, enabling more efficient information acquisition.
Similarly, removing usage history does not significantly degrade accuracy but increases the number of queries. 
This indicates that usage history serves as a complementary signal that reduces uncertainty and improves interaction efficiency.
Overall, these results show that active questioning is the most critical component for achieving high accuracy, while user background information and usage history help reduce the number of required queries.




Figure~\ref{fig:result_ablation_step} shows the mean and standard deviation of Subset Accuracy at each query step over 10 trials for the ablation settings with active questioning.
When user background information is removed, the model achieves slightly higher Subset Accuracy in the early stages (up to Step 2) than the full model. This suggests that simplifying the input information can make initial predictions easier by reducing ambiguity.
However, as interaction proceeds, the proposed method consistently outperforms this variant. This indicates that the absence of background information limits the model's ability to incorporate the contextual cues needed to refine ownership estimates.
These results show that user background information is not critical for initial predictions but is important for sustaining performance improvements and achieving accurate ownership estimation in later stages.



Next, we analyze the effect of removing usage history.
Although the final performance of this variant is comparable to the proposed method, its performance at each query step is consistently lower throughout the inference process. In addition, more query steps are required to approach 1.0 in Subset Accuracy. These results indicate that the absence of usage history leads to higher uncertainty in the early stages, requiring additional interactions to resolve ambiguity.

In contrast, the proposed method achieves higher accuracy early on by leveraging usage history as a strong prior for ownership estimation. 
However, as interaction proceeds, newly acquired information -- such as temporary usage or exceptional behaviors -- is incorporated through user responses, which can temporarily destabilize the estimation. 
Despite this, the uncertainty-aware querying mechanism gradually resolves these inconsistencies, leading to convergence toward the correct ownership set.
Overall, these results suggest that usage history primarily reduces early-stage uncertainty and improves query efficiency, while active questioning ensures final accuracy.




\section{Limitation}\label{sec:limitation}

This study has several limitations.
First, the proposed method relies heavily on an LLM for ownership inference. 
While the LLM can flexibly integrate diverse contextual information such as user background and usage history, its outputs are not guaranteed to be stable or reproducible. The estimation results may vary depending on prompt design and input representation, and the reasoning process is less interpretable than that of structured probabilistic models.

Second, the performance of the proposed method strongly depends on the quality and accuracy of contextual information, including user background and object usage history. 
In this study, we assume that such information is reliably available. 
However, in real-world environments, usage history may be incomplete or noisy, and background information may be inaccurate. In such cases, the LLM's inference based on commonsense reasoning may not align with the true ownership.
Addressing robustness under imperfect or unreliable contextual information remains an important direction for future work.



\section{Conclusion}\label{sec:conclusion}

In this paper, we proposed a context-aware and uncertainty-driven interactive method for object ownership estimation in everyday environments. 
The proposed method integrates user background information and object usage history as contextual cues, and leverages an LLM to infer ownership. 
By incorporating Conformal Prediction (CP), the method selectively generates questions only when the estimation is uncertain, enabling efficient information acquisition.
Experimental results demonstrated that the proposed method can accurately estimate ownership sets by combining contextual inference with uncertainty-driven interaction. 
In particular, the method effectively utilizes contextual information for initial estimation and progressively refines ownership predictions through user interaction, achieving high accuracy while reducing unnecessary queries.

For future work, extending the framework to explicitly model the temporal dynamics of ownership remains an important direction. 
While this study considers short-term usage patterns such as temporary borrowing, it does not fully address long-term ownership changes or role transitions. 
Modeling the temporal evolution of ownership scores would enable more robust handling of dynamic environments.
Another important direction is to handle open-world settings where new users and objects are introduced. 
Incorporating mechanisms to adaptively integrate unseen users into the ownership inference process, while maintaining efficient interaction, will be essential for real-world deployment.
These extensions would further enhance the applicability of the proposed method to long-term and real-world human–robot interaction scenarios.




\section*{Author Contributions}
\textbf{Saki Hashimoto}: Experiments and Evaluation, Writing – Original Draft, Writing – Review \& Editing.  
\textbf{Akira Taniguchi}: Conceptualization, Writing – Original Draft, Writing – Review \& Editing, Supervision, Funding Acquisition.  
\textbf{Shoichi Hasegawa}: Writing – Original Draft, Writing – Review \& Editing.  
\textbf{Yoshinobu Hagiwara}: Writing – Original Draft, Writing – Review \& Editing.  
\textbf{Tadahiro Taniguchi}: Conceptualization, Writing – Original Draft, Writing – Review \& Editing, Supervision, Funding Acquisition.  






\section*{Funding}
This study was supported by 
JST Moonshot R\&D Program, grant number JPMJMS2011; and
JSPS KAKENHI, grant numbers JP23K16975 and 	JP26K14952, Japan.

\bibliographystyle{tfnlm}
\bibliography{article}

\appendix
\section{Algorithm}\label{app:algorithm}
The overall procedure is summarized in Algorithm~\ref{alg:ownership_single_shot}.
The main components of the proposed method are defined as follows. 
\textbf{BUILD-KNOWN} extracts reliable ownership information from the estimation state $D$ and constructs a set of known facts $K$. 
\textbf{LLM-INFER} estimates multi-label ownership scores for each candidate user based on object class, neighboring objects, similar objects, and usage history. 
\textbf{DETECT-SHARE} determines whether an object is likely to be shared based on the estimated ownership scores. 
\textbf{UNCERTAINTY} quantifies estimation uncertainty using the size of the prediction set $\Gamma(x)$ obtained by CP. 
\textbf{QUESTION-GENERATION} generates natural language questions for ownership confirmation based on the current estimation state $D[o^\star]$. 
\textbf{UNDERSTAND-ANSWER} interprets the user's response using the LLM and converts it into deterministic ownership scores. 
\textbf{APPLY-ANSWER} updates the ownership estimation state $D$ using the acquired information and marks the object as queried.

\begin{figure}[!t]
\begin{algorithm}[H]
    \small
    \caption{Active Ownership Acquisition via Single-shot LLM Inference}
    \label{alg:ownership_single_shot}
    \begin{algorithmic}[1]
        \STATE \textbf{Input:}
        object set $\mathcal{O}$,
        user set $\mathcal{U}$,
        coverage threshold $q_{\alpha}$,
        stopping threshold $q_{\mathrm{cp}}$,
        maximum number of queries $Q_{\max}$

        \STATE \textbf{Initialize:}
        ownership estimation state $D[o]$ for all $o \in \mathcal{O}$,
        query counter $q_{\textrm{cnt}} \gets 0$

        \WHILE{true}
            \STATE $K \gets \textbf{BUILD-KNOWN}(D)$

            \FORALL{objects $o \in \mathcal{O}$ with $D[o].asked = 0$}
                \STATE $\mathcal{I}(o) \gets
                \langle \text{class}(o), \mathcal{N}(o), \mathcal{S}(o), \mathcal{U}(o) \rangle$

                \STATE $\mathbf{s}(o)
                \gets \textbf{LLM-INFER}(\mathcal{I}(o))$

                \STATE $(k_{\mathrm{share}}, M_{\mathrm{share}})
                \gets \textbf{DETECT-SHARE}(\mathbf{s}(o))$

                \STATE $D[o] \gets
                \langle \mathbf{s}(o), k_{\mathrm{share}}, M_{\mathrm{share}} \rangle$
            \ENDFOR

            \STATE $\Gamma(o) \gets
            \{u \in \mathcal{U} \mid s_o(u) \ge 1-q_{\alpha}\}$

            \STATE $\mathrm{cp}_s(o) \gets \textbf{UNCERTAINTY}(\Gamma(o))$

            \IF{all $\mathrm{cp}_s(o) \le q_{\mathrm{cp}}$}
                \STATE \textbf{break}

            \ENDIF

            \STATE $o^\star \gets \arg\max_o \mathrm{cp}_s(o)$

            \STATE $q_{o^\star}
            \gets \textbf{QUESTION-GENERATION}(D[o^\star])$

            \STATE $w_{o^\star}
            \gets \textbf{UNDERSTAND-ANSWER}(o^\star)$

            \STATE $D \gets \textbf{APPLY-ANSWER}(D, o^\star, w_{o^\star})$

            \STATE $q_{\textrm{cnt}} \gets q_{\textrm{cnt}} + 1$
            \IF{$q_{\textrm{cnt}} \ge Q_{\max}$}
                \STATE \textbf{break}
            \ENDIF
        \ENDWHILE
    \end{algorithmic}
\end{algorithm}
\end{figure}

\section{NLMap}\label{app:NLMap}

Natural Language Map (NLMap)~\cite{NLMap} is an open-vocabulary semantic mapping framework that enables robots to reference objects in real environments using natural language. The term `open-vocabulary' indicates that object categories are not fixed in advance, allowing flexible reference expressions beyond predefined labels.

Unlike conventional semantic maps that rely on a fixed set of object classes (\textit{e.g.}, \textit{cup}, \textit{bottle}), NLMap represents objects using visual-language features, enabling retrieval based on arbitrary natural language descriptions (\textit{e.g.}, `a cup with a handle' or `a blue container').

\textbf{Map Representation}
During environment exploration, object-like regions are detected from RGB-D observations without restricting object categories. For each detected region $I_i$, a visual feature $\phi_i$ is extracted using vision-language models such as CLIP~\cite{CLIP} and ViLD~\cite{ViLD}. In addition, the 3D position $p_i$ and object size $r_i$ are estimated from depth information.

Each object candidate is represented as a context element:
\begin{equation}
c_i = (\phi_i, p_i, r_i),
\end{equation}
and the environment is represented as a set of context elements:
\begin{equation}
C = \{c_i\}_{i=1}^{N}.
\end{equation}

\textbf{Language-based Retrieval}
Given a natural language query, a text feature is obtained using a text encoder, and object retrieval is performed based on similarity with context elements. A similarity function is defined as:
\begin{equation}
D : C \times Y \rightarrow \mathbb{R},
\end{equation}
$Y$ is the set of object names.
The similarity is computed as the maximum of CLIP-based and ViLD-based similarities between visual and textual features. This design balances robustness to known objects with generalization to unseen ones.

Relevant objects are retrieved using top-$k$ nearest neighbor search based on $D$, followed by thresholding to determine object presence.

\textbf{Multi-view Integration}
Since the same object may be observed from multiple viewpoints, NLMap integrates multiple context elements corresponding to a single object instance. Each element is modeled as a Gaussian distribution:
\begin{equation}
G_i = \mathcal{N}(p_i,\; \alpha r_i I),
\end{equation}
and candidate elements are merged based on the KL divergence:
\begin{equation}
\mathrm{KL}(G_i \| G_j).
\end{equation}
Elements with divergence below a threshold are considered to represent the same object and are merged accordingly.

\textbf{Summary}
NLMap provides an open-vocabulary representation that combines visual-language features with geometric information, enabling flexible, robust object retrieval via natural-language queries in real-world environments.

\section{Conformal Prediction}\label{sec:CP}

We introduce Conformal Prediction (CP)~\cite{CP2005} to quantify uncertainty in ownership estimation, rather than directly interpreting the LLM's ownership scores as probabilities. Specifically, CP constructs a prediction set of candidate owners, and the size of this set is used as an uncertainty measure.

CP is a framework for uncertainty quantification that provides statistical guarantees without requiring distributional assumptions. Given a user-specified error rate $\alpha$, CP constructs a prediction set that contains the true label with probability at least $1 - \alpha$, under the assumption of exchangeable data~\cite{CP2005,CP2023}.

\textbf{Basic Procedure}
Let $\hat{f}$ be a predictive model that outputs scores for input $x$. Given a calibration dataset $\{(X_i, Y_i)\}_{i=1}^n$, a nonconformity score $s(X_i, Y_i)$ is defined to measure the discrepancy between the prediction and the true label. Larger values indicate lower confidence.

A quantile $\hat{q}$ corresponding to the error level $\alpha$ is then computed from the calibration scores. At test time, the prediction set is defined as:
\begin{align}
C(X_{\mathrm{test}}) = \{y \mid s(X_{\mathrm{test}}, y) \le \hat{q}\}.
\end{align}

This construction satisfies the following guarantee:
\begin{align}
\Pr\bigl(Y_{\mathrm{test}} \in C(X_{\mathrm{test}})\bigr) \ge 1 - \alpha.
\end{align}

\textbf{Interpretation}
The output of CP is a set of candidate labels rather than a single prediction. The size of the prediction set reflects the model's uncertainty: smaller sets indicate more confident predictions, while larger sets indicate more ambiguous or uncertain cases.

\textbf{Application in This Study}
In this work, the LLM's ownership scores are treated as heuristic outputs, and CP is applied to construct prediction sets for candidate owners. The number of users included in the prediction set is used as an uncertainty measure.

Objects with larger prediction sets are considered more uncertain and are prioritized for question generation. In this way, CP provides both a statistically grounded interpretation of uncertainty and a principled mechanism for selecting informative queries.

\section{Model Context Protocol}\label{sec:MCP}

We use the Model Context Protocol (MCP) to provide object usage history in a structured, on-demand manner for ownership inference.

Rather than directly including raw usage history (\textit{e.g.}, natural language captions) in the prompt, we encapsulate the aggregation and summarization process as an external tool. This allows the LLM to retrieve only essential information, such as user-specific usage frequency, action statistics, and recent interactions, when needed, avoiding prompt bloat and improving the clarity of contextual evidence.

MCP is an open protocol proposed by Anthropic that enables standardized interaction between AI models and external tools or data sources~\cite{anthropic_mcp}. It consists of three main components: Host, Client, and Server~\cite{MCP2025}. The Host (\textit{e.g.}, a chat application) manages the interaction, the Client mediates communication, and the Server provides access to external functionalities.

MCP servers expose three types of capabilities: (i) \textit{tools}, (ii) \textit{resources}, and (iii) \textit{prompts}. Tools enable execution of external functions, resources provide access to data sources, and prompts define reusable templates. This unified framework allows both data retrieval and action execution to be handled consistently.

\textbf{Application in This Study}
In our framework, usage history is processed by an MCP server that summarizes interaction logs into structured representations. When needed, the LLM invokes this tool to obtain concise and relevant usage information for a target object.

This design offers two key advantages. First, it prevents excessive input length and preserves the salience of critical information for inference. Second, it separates data preprocessing from LLM reasoning, improving stability and reproducibility of the ownership estimation process.

As a result, MCP enables efficient integration of rich contextual information while maintaining scalable and robust inference.






\section{Dataset Generation Details}\label{app:dataset_generation}

This section describes the detailed procedure for constructing the usage history dataset from the HOMER dataset.

\subsection{Event Construction}

We sample $N=7$ days of activity schedules and expand them into sequences of interaction events. Each event is represented as $e_i = (d_i, t_i, u_i, a_i, o_i)$,
where $d_i$, $t_i$, $u_i$, $a_i$, and $o_i$ denote the date, time, user, action, and object, respectively.

Original action labels (\textit{e.g.}, \texttt{GRAB}, \texttt{USE}, \texttt{PUTBACK}) are converted into natural language descriptions. These events are sorted chronologically to form an event sequence $E = \{ e_1, e_2, \dots, e_T \}$.

\subsection{Usage History Construction}

For each object $o$, we construct a usage history $H(o) = \langle h_1, h_2, \dots, h_{n_o} \rangle$, where each entry $h_j = (d_j, t_j, \text{text}_j)$ represents a natural language interaction event.

\subsection{Ownership Assignment}

We define the set of true owners for object $o$ as 
$O(o) = \{u_1, u_2, \dots\}$.
Three usage scenarios are considered:

\textbf{Single-user}: 
Only the owner $u^\ast$ interacts with the object:
\[
P(u \mid o) =
\begin{cases}
1 & (u = u^\ast) \\
0 & \text{otherwise}
\end{cases}
\]

\textbf{Temporary Sharing}:
We introduce a borrowing probability $p_{\mathrm{borrow}}(o)$:
\[
P(u \mid o) =
\begin{cases}
1 - p_{\mathrm{borrow}}(o) & (u \in O(o)) \\
\dfrac{p_{\mathrm{borrow}}(o)}{|U \setminus O(o)|} & (u \notin O(o))
\end{cases}
\]

\textbf{Multi-user Sharing}:
Users are selected based on usage balancing:
\[
u_k = \arg\min_{u \in O(o)} c(u,o),
\]
where $c(u,o)$ is the cumulative usage count.

\subsection{Session Construction}

Events are grouped into sessions $S(o) = \{ s_1, s_2, \dots, s_{m_o} \}$, where each session corresponds to consecutive interactions by a single user. If explicit interaction boundaries are unavailable, sessions are segmented based on a time gap threshold $\Delta t$.

Each session is assigned a single user:
\[
P(u_i = u_k \mid e_i \in s_k) = 1.
\]

\subsection{Data Split}

The dataset is divided chronologically. Let $t_{\min}$ be the earliest timestamp. We define $t_{\mathrm{cut}} = t_{\min} + 3\ \mathrm{days}$.
Events before $t_{\mathrm{cut}}$ are used for training, and the rest for evaluation.

\section{Evaluation Metrics}\label{app:metrics}

This section provides the formal definitions of the evaluation metrics used in this study.

\subsection{Subset Accuracy}

Subset Accuracy measures the proportion of samples for which the predicted ownership set exactly matches the ground-truth set:
\begin{equation}
\mathrm{SubsetAccuracy}
=
\frac{1}{N}
\sum_{i=1}^{N}
\mathbb{I}(U_i = \hat{U}_i)
\end{equation}

This metric is used to evaluate whether the ownership set is estimated exactly correctly.  
Let $N$ denote the number of samples, $U_i$ the ground-truth owner set, and $\hat{U}_i$ the predicted owner set.

\subsection{Mean Jaccard}

The Jaccard index evaluates partial overlap:
\begin{equation}
\mathrm{Jaccard}(U_i, \hat{U}_i)
=
\frac{|U_i \cap \hat{U}_i|}{|U_i \cup \hat{U}_i|}
\end{equation}

\begin{equation}
\mathrm{MeanJaccard}
=
\frac{1}{N}
\sum_{i=1}^{N}
\mathrm{Jaccard}(U_i, \hat{U}_i)
\end{equation}

\subsection{Micro Precision, Recall, and F1}

We compute micro-averaged precision and recall over all samples and users:
\begin{equation}
\mathrm{MicroPrecision}
=
\frac{\sum_{i,u} \mathrm{TP}_{i,u}}
     {\sum_{i,u} (\mathrm{TP}_{i,u} + \mathrm{FP}_{i,u})}
\end{equation}

\begin{equation}
\mathrm{MicroRecall}
=
\frac{\sum_{i,u} \mathrm{TP}_{i,u}}
     {\sum_{i,u} (\mathrm{TP}_{i,u} + \mathrm{FN}_{i,u})}
\end{equation}

\begin{equation}
\mathrm{MicroF1}
=
\frac{2 \cdot \mathrm{MicroPrecision} \cdot \mathrm{MicroRecall}}
     {\mathrm{MicroPrecision} + \mathrm{MicroRecall}}
\end{equation}

Here, $\mathrm{TP}_{i,u}$ (True Positive) indicates that user $u$ is correctly predicted as an owner of object $i$, \textit{i.e.}, $u \in U_i \cap \hat{U}_i$.
$\mathrm{FP}_{i,u}$ (False Positive) indicates that user $u$ is predicted as an owner but is not a true owner, \textit{i.e.}, $u \in \hat{U}_i \setminus U_i$.
$\mathrm{FN}_{i,u}$ (False Negative) indicates that user $u$ is a true owner but is not predicted as such, \textit{i.e.}, $u \in U_i \setminus \hat{U}_i$.

\subsection{Number of Questions}

The number of questions is defined as the total number of queries issued by the robot until ownership is determined.

\section{Analysis of Query Target Selection}\label{app:query_analysis}

\begin{figure}[tb]
    \centering
    \includegraphics[width=0.6\linewidth]{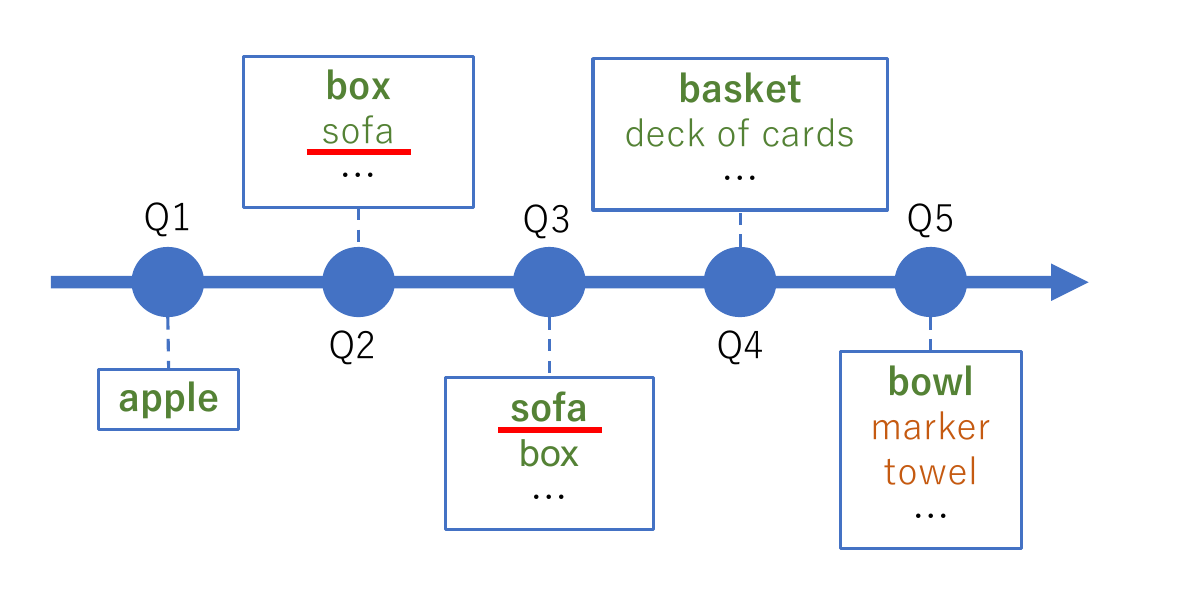}
    \caption{
    Examples of objects selected for querying at each step across 10 trials.
    Each $Q_k$ denotes the $k$-th query step. Blue boxes indicate the objects selected for querying, while red underlines denote objects whose ownership can be inferred from their class labels.
    }
    \label{fig:queried_objects}
\end{figure}

Figure~\ref{fig:queried_objects} shows examples of objects selected for questioning at each step.

The proposed method tends to prioritize objects with high uncertainty, particularly those associated with multiple candidate owners. This behavior arises from the use of conformal prediction, where objects with larger prediction sets are more likely to be selected for querying.

However, the method also selects objects that are intuitively shared (\textit{e.g.}, sofa, bed), even when their ownership could be inferred from commonsense knowledge. This suggests that such prior knowledge is not fully utilized in the current uncertainty estimation process.

These observations highlight a limitation of the proposed approach and suggest that incorporating stronger priors or commonsense reasoning into uncertainty estimation is a promising direction for future work.

\end{document}